\documentclass[lettersize,journal]{IEEEtran}
\usepackage[dvipsnames]{xcolor}
\usepackage{graphicx}
\usepackage{subcaption}
\usepackage{tabularx}
\usepackage{amsfonts} 
\usepackage{graphicx}
\usepackage{amsmath}
\usepackage{comment}
\usepackage{biblatex}
\usepackage{hyperref}
\usepackage{xcolor}
\newtheorem{prop}{Proposition}

\usepackage[ruled,vlined]{algorithm2e}
\usepackage{algorithmicx}
\usepackage{algpseudocode}

\usepackage{lipsum}
\usepackage[dvipsnames]{xcolor}
\usepackage{color}
\begin{document}

\title{RF-Enhanced Road Infrastructure for \\ Intelligent Transportation}

\author{Dajiang Suo$^{1,3}$, Heyi Li$^{1}$, Rahul Bhattacharyya$^{1}$, Zijin Wang$^{2}$, Shengxuan Ding$^{2}$, Ou Zheng$^{2}$, \\ Daniel Valderas$^{4}$, Joan~Meli\`{a}-Segu\'{i}$^{5}$, Mohamed Abdel-Aty$^{2}$, Sanjay E. Sarma$^{1}$
        % <-this % stops a space

\thanks{$^{1}$The authors are with the Department of Mechanical Engineering,
        Massachusetts Institute of Technology, Cambridge, MA 02139, USA
        {\tt\small djsuo,heyi,rahul$\_$b,sesarma@mit.edu}}% <-this % stops a space
\thanks{$^{2}$The authors are with the Department of Civil, Environmental and Construction Engineering,
        University of Central Florida, Orlando, FL, 32816, USA
        {\tt\small zijin.wang,shengxuan.ding ou.zheng ,m.aty\\@ucf.edu}}%
\thanks{$^{3}$The author is with the Polytechnic School in the Ira A. Fulton Schools of Engineering, Arizona State University, Tempe, AZ 85281, USA
        {\tt\small dajiang.suo@asu.edu}}%
\thanks{$^{4}$The author is with the Tecnun School of Engineering, University of Navarra, San Sebastián, Spain
        {\tt\small dvalderas@tecnun.es}}%
\thanks{$^{5}$The author is with the Faculty of Comp. Science, Multimedia and Telecom., Universitat Oberta de Catalunya (UOC), Barcelona, Spain
        {\tt\small melia@uoc.edu}}%
\thanks{Corresponding author: Dajiang Suo (dajiang.suo@asu.edu)}

}

% The paper headers
\markboth{}%
{Shell \MakeLowercase{\textit{et al.}}: A Sample Article Using IEEEtran.cls for IEEE Journals}

%\IEEEpubid{0000--0000/00\$00.00~\copyright~2021 IEEE}
% Remember, if you use this you must call \IEEEpubidadjcol in the second
% column for its text to clear the IEEEpubid mark.

\maketitle

\begin{abstract}
The EPC GEN 2 communication protocol for Ultra-high frequency Radio Frequency Identification (RFID) has offered a promising avenue for advancing the intelligence of transportation infrastructure. With the capability of linking vehicles to RFID readers to crowdsource information from RFID tags on road infrastructures, the RF-enhanced road infrastructure (REI) can potentially transform data acquisition for urban transportation. Despite its potential, the broader adoption of RFID technologies in building intelligent roads has been limited by a deficiency in understanding how the GEN 2 protocol impacts system performance under different transportation settings. This paper fills this knowledge gap by presenting the system architecture and detailing the design challenges associated with REI. Comprehensive real-world experiments are conducted to assess REI's effectiveness across various urban contexts. The results yield crucial insights into the optimal design of on-vehicle RFID readers and on-road RFID tags, considering the constraints imposed by vehicle dynamics, road geometries, and tag placements. With the optimized designs of encoding schemes for reader-tag communication and on-vehicle antennas, REI is able to fulfill the requirements of traffic sign inventory management and environmental monitoring while falling short of catering to the demand for high-speed navigation. In particular, the Miller 2 encoding scheme strikes the best balance between reading performance (e.g., throughput) and noise tolerance for the multipath effect. Additionally, we show that the on-vehicle antenna should be oriented to maximize the available time for reading on-road tags, although it may reduce the received power by the tags in the forward link. 
\end{abstract}


\section{Introduction}

\IEEEPARstart{T}he increasing complexity of urban transportation necessitates technological solutions that can significantly improve data gathering and consequently inform decision-making processes in traffic control and transportation management. Government bodies like the European Commission are currently defining rules to boost intelligent transport systems for safer and more efficient transportation, including road digitization. One such technology is RF-enhanced road infrastructure (REI), a system that deploys RFID tags on roads to communicate with on-vehicle readers. This allows for a comprehensive, localized collection of road and environmental data with significant potential for crowdsourcing. Existing research validates the possibility of utilizing RFID tags, in line with the EPC GEN 2 protocol~\cite{epcglobal2013epc}, to aid vehicle navigation, traffic sign recognition, and environmental condition monitoring. This protocol facilitates communication between the interrogator and passive RFID tags that operate within the ultra-high frequency (UHF) range of 860 - 960 MHz.

Despite its promise, the wide adoption of REI has been hindered by the lack of understanding of the implications of EPC GEN 2 protocol for the system performance metrics under different vehicle dynamics and transportation contexts. To maintain system stability, the positioning and configuration of the REI system should be adapted to different configurations of on-board systems, vehicle speeds, road geometries, and the placement of tags on roads.

This paper addresses the knowledge gaps in REI designs by examining the effect of the EPC UHF GEN 2 protocol within our proposed REI system. Comprehensive experimental evaluations of REI performance are carried out on real-world urban streets to investigate the system under practical conditions. The preliminary results provide insights into the optimal design of on-vehicle RFID readers and on-road RFID tags, considering the influences of road geometries and the constraints imposed by the tag-reader communication protocol (EPC GEN 2). 

In particular, with the optimized designs of encoding schemes for reader-tag communication and on-vehicle antennas, REI can fulfill the requirements of traffic sign inventory management and environmental monitoring while falling short of catering to the demand for high-speed navigation. In particular, the Miller 2 encoding scheme strikes the best balance between reading performance (e.g., throughput) and noise tolerance for the multipath effect. Additionally, we show that the on-vehicle antenna should be oriented to maximize the available time for reading on-road tags, although it may reduce the received power by the tags in the forward link. Key technologies that enhance the wide adoption of REI in intelligent transportation are also presented.

The remainder of this paper is organized as follows. Section II presents related work on RFID usage in transportation contexts. In Section III, we present the high-level system architecture for REI. The detailed use scenarios and preliminary designs are given in Section IV. In Section V, we perform performance analyses on REI and discuss the influencing factors based on experiment results from real-world tests. Section VI discusses technology gaps in using current UHF GEN II RIFD to support transportation applications. Section IX concludes by summarizing the main results. 

\section{Related work}
RFID has been used in various transportation applications, including vehicle localization and navigation~\cite{khosyi2020tests,meghana2017comprehensive,naik2018rfid,qin2021collision,wang2014rfid,suo2023rf}. Conventional localizing approaches use satellite navigation systems \cite{shi2022beyond}, vehicle motion sensors \cite{gao2022improved,trogh2020map}, range sensors \cite{burnett2022we}, and vision sensors \cite{gurghian2016deeplanes,tarel2009fast}. For relative localization, visual odometry is the dominant approach. However, the accuracy of localization may be substantially impacted under conditions of adverse weather and low visibility \cite{rasshofer2005automotive}, where RFID technology can assume an essential role. In particular, RFID tags, when being deployed on the surface of roads, can serve the roles of pavement markings for lane detection and keeping~\cite{cheng2011design,javaid2021autonomous,kawamura2014intelligent,zheng2016lane}.

Another type of RFID application in transportation contexts involves inventory management of transportation assets. To maintain, upgrade, and operate transportation assets such as traffic signs, the transportation department expends a significant amount of effort on manual checking and updating every year. The recent emergence of semi-automatic management approaches has enhanced management efficiency and reduced human labor \cite{ellison2008tapping}. Applying computer vision technologies for sign detection has been widely studied and adopted \cite{tabernik2019deep,maldonado2008traffic}, however, it still face challenges in nighttime detection and occlusion conditions. Other approaches use communication technologies such as DSRC and Zigbee to overcome the constraints of vision-based method \cite{li2009traffic,wang2021traffic,garcia2011robust}. A significant limitation of this approach is the energy consumption, since continuous power supply is needed for traffic signs This necessitates additional effort for managing and maintaining the battery. Therefore, RFID, as a passive detection method, perfectly addresses this issue. For example, RFID tags are installed on traffic signs such that any vehicle with RFID readers can interrogate sign status and update the cloud server as the vehicle moves~\cite{chen2022rfid,garcia2018passive,hidalgo2013wireless}.

Additionally, since RFID can be combined with various types of sensors for environmental monitoring in supply chains and agriculture~\cite{mezzanotte2021innovative,palazzi2019feeding}, it can also be used to fulfill the need for real-time monitoring of road and weather conditions in transportation contexts~\cite{costa2021review}.
 
\section{RFID On Roads}
\subsection{System architecture}

\begin{figure*}[tb!]
\centering
\includegraphics[width=0.8\textwidth]{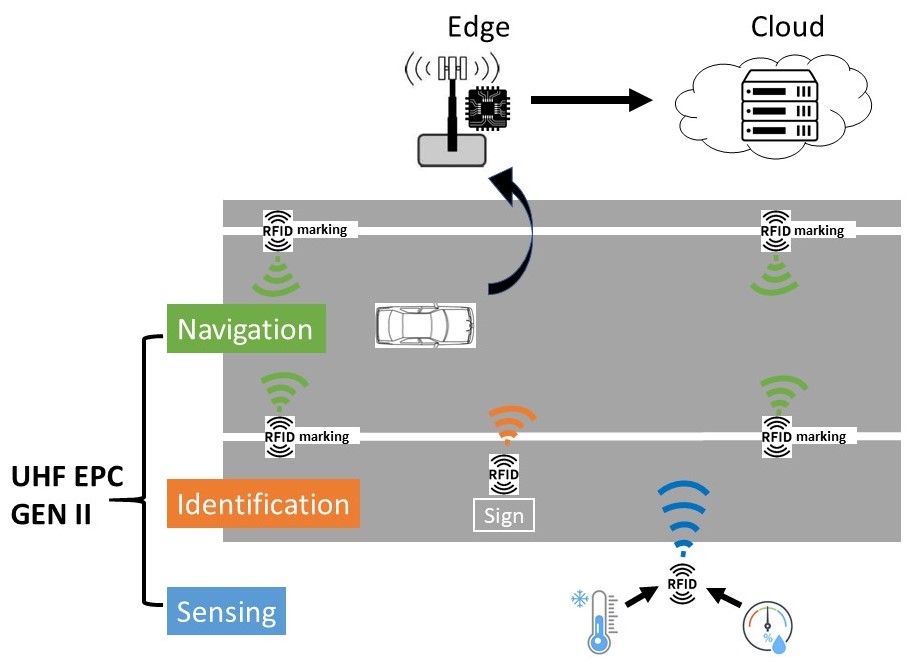}
\caption{The architecture of RFID on Roads System}
\label{fig:sys_arch}
\end{figure*}

\begin{figure*}[tb!]
\centering
\includegraphics[width=0.95\textwidth]{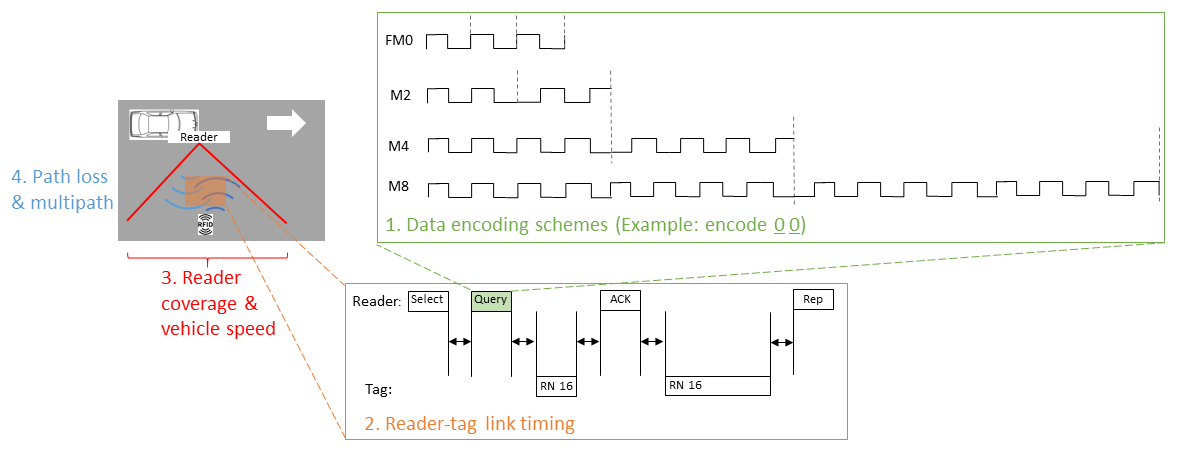}
\caption{The influencing factors for the performance of RF-enhanced road infrastructure.}
\label{fig:performance_fact}
\end{figure*}

The architecture of the REI, as depicted in Fig.~\ref{fig:sys_arch}, revolves around deploying RFID tags on roads. These tags store information about roads, signs, and environmental conditions. For UHF tags compliant with the EPC GEN 2 protocol, such information may be encoded within the EPC or user memory bank. Vehicles equipped with RFID readers can then make inquiries to these on-road tags as they traverse, thereby collecting road and environment-related data in real-time. This data can support vehicle navigation, road object identification, and environmental sensing.

When on-road tags are encoded with information that corresponds to pavement markings, they can provide support to vehicle navigation. For example, the tags that are placed next to lane markings can inform nearby vehicles about the lane boundaries, driving directions, and speed limits. This provides opportunities for lane-level positioning and automated vehicle navigation under adverse environmental conditions.

Similarly, the tags that are attached to traffic signs may be used to identify their identities and types. In addition to navigation and identification, roadside tags may be used to realize sensing functionalities. Temperature and humidity data collected through RFID tag-powered sensors, for example, may be stored within tag memory and interrogated by onboard readers when vehicles pass by.

\subsection{Design considerations and challenges}

Each on-road tag must be successfully read at least once by a moving vehicle to avoid missing crucial data. However, meeting this requirement poses a challenge due to several factors:

\begin{itemize}
    \item Path loss or interference. Our REI utilizes passive RFID, which leverages backscatter communication for data exchange. Consequently, the strength of the RF signal received by tags decreases in proportion to the distance squared. Therefore, REI design needs to factor in various distances between on-vehicle readers and tags under different road geometries. Moreover, environmental noise resulting from the multi-path effect can destabilize signal strength and degrade system performance.
    \item Limited reader coverage. Every onboard reader on a moving vehicle has limited opportunities to interrogate each tag and receive responses. As the vehicle speed increases, the time duration for completing the reading event diminishes.
    \item Reader-tag command string sequences. The communication between a reader and a tag follows the EPC GEN 2 protocol, which necessitates a specific amount of time to complete the data exchange cycle or 'inventory round'. Command strings such as RN16 and ACK must occur in sequence and meet the timing requirements stipulated by EPC GEN 2. For instance, after a tag backscatters an RN16 (a 16-bit random number), the reader must respond with an ACK command within a pre-defined duration. Failure to do so results in a failed interrogation, necessitating the restart of the inventory round. These failures can arise due to environmental noise or loss of tag power.
    \item Data encoding schemes. Beyond command sequences, the data encoding schemes used for backscattered communication also affect the link timing and thus, the success rate of inventory rounds. UHF tags may use either Frequency Modulation Zero (FM0) or Miller Modulation. Although FM0 typically requires more bandwidth but less time to transmit the same amount of data, making it beneficial for moving vehicles with limited reader coverage, Miller Modulation offers greater resistance to noise. This quality is crucial for REI applications in outdoor environments where multi-path effects can cause interference.  \item Power Sensitivity. A minimum amount of power is required to establish the communication for both tags and readers. The power sensitivity denotes the minimum power so that tags could decode commands and reply, and readers could decode response from tags. While car is moving, power gain is altering, as radiation pattern is changing with relative angles, speed and locations between readers and tags. 
\end{itemize}

\section{Use cases and preliminary designs}
\subsection{RF pavement markings}
\paragraph{Motivation}
Traditional lane detection algorithms that use information from visual channels can falter in adverse weather conditions and situations of low visibility. As a solution to these challenges, we propose the implementation of a Radio Frequency (RF) pavement marking system. This system leverages Radio Frequency Identification (RFID) and is proposed as a supplementary system to existing computer vision-based approaches. The primary goal of this RFID-based system is to enhance lane detection accuracy and support the autonomous operation of vehicles under challenging environmental conditions.

\paragraph{Preliminary designs and challenges}
\begin{figure}[tb!]
\centerline{\includegraphics[width=0.35\textwidth]{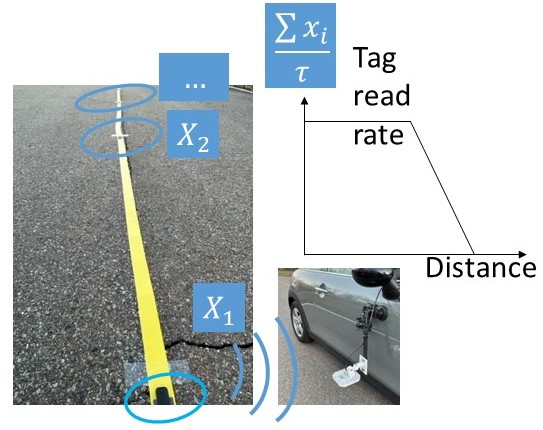}}
\caption{The theoretical relation between RFID read rate and the distance between the antenna and the tag~\cite{suo2023rf}.}
\label{fig:filtering}
\end{figure} 

\begin{figure}[tb!]
\centerline{\includegraphics[width=0.4\textwidth]{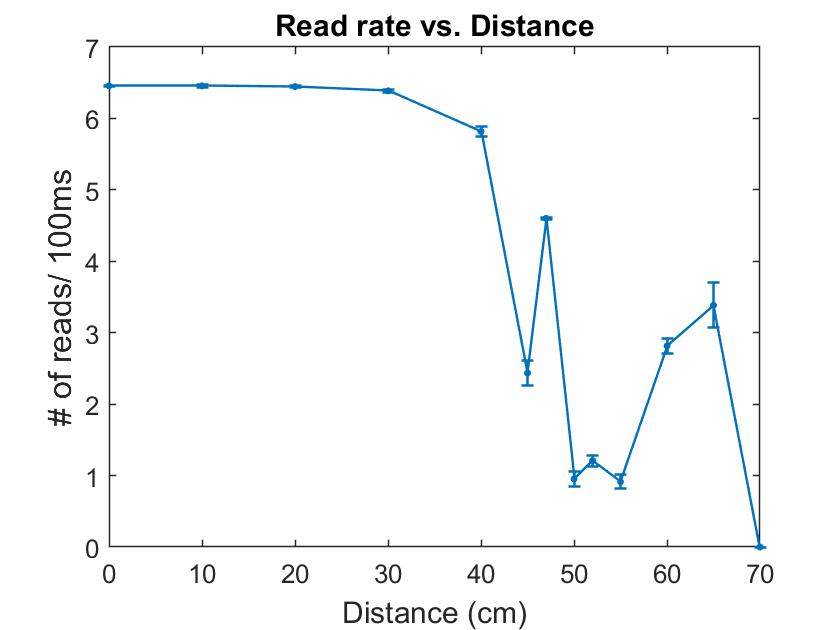}}
\caption{The number of tag readings as the tag moves away from the reader antenna~\cite{suo2023rf}.}
\label{fig:readrate}
\end{figure}

In an ideal condition, when a vehicle is stationary in the longitudinal direction of the lane, the read rate of individual RFID tags—defined as the frequency of tag detection by the on-vehicle antenna within a given time interval—decreases as the RFID tag moves laterally away from the antenna (Fig.~\ref{fig:filtering}). However, using this method for precise lane-level vehicle localization is challenging due to noise from multi-path effects and the influence of the vehicle's longitudinal and lateral motion on signal properties. Consequently, the read rate becomes an unreliable distance indicator between the vehicle and the lane edge.

To validate this assumption, we measured the number of readings of a single tag within a duration of 100 milliseconds using the RFID reader and tags from our experimental setup. Fig.~\ref{fig:readrate} presents the results obtained from ten repetitions of the experiment conducted on concrete roads under dry conditions. Indeed, as the tag moves away from the center of the on-vehicle reader antenna, the tag read rate initially decreases, fluctuates, and ultimately approximates zero.

In our previous work~\cite{suo2023rf}, we present an RF lane marking system, which uses read rate to estimate the target vehicle's relative position to the lane. To overcome the influence of the multi-path effect and vehicle motion on RFID reading events, a probabilistic filtering algorithm rooted in sampling theory is proposed. This algorithm utilizes a Poisson Bernoulli distribution to reduce variance in tag-reading probability due to environmental noises.

Specifically, we form a random variable Z, by summing n individual random variables $X_i$, where n is the number of interrogations the RFID reader makes within a user-defined interval $\tau$, and $X_i \sim Bernoulli(p_i)$ denotes whether the on-vehicle antenna successfully detects a given tag in the $i^{th}$ time slot within that interval. 
\begin{center}
$Z = \sum_{i=1}^n X_i$
\end{center}  
Z follows a Poisson Bernoulli distribution, which allows us to infer the distance between the vehicle and the lane edge. 

\begin{figure}[tb!]
\centerline{\includegraphics[width=0.35\textwidth]{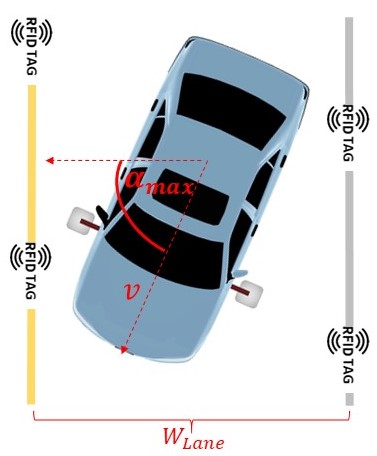}}
\caption{The constraint on the maximum value of time $\tau$ imposed by the safety requirement of departure warning~\cite{suo2023rf}.}
\label{fig:safety_constraint}
\end{figure}

\begin{prop}\label{prop_tau_min}
Increasing the value of the user-defined interval $\tau$ can enhance the reliability of the tag read rate as an indicator of the lateral distance between the on-vehicle antenna and the lane.
\end{prop}

By observing the variance of sample mean $\overline{X}=\frac{Z}{n}$ and assuming independent reading events, by sampling theory, we have
\begin{center}
$Var\overline{X} = \frac{1}{n^2}\sum_{i=1}^nVar(X_i)=\frac{\sigma^2}{n} $
\end{center}
This suggests that opting for a larger user-defined interval $\tau$ helps mitigate the impact of RF reading noise. However, it is essential to consider safety requirements when determining the upper limit of $\tau$.

\begin{prop}\label{prop_max_tau}
The user-defined time interval $\tau$ for updating lane-tag readings must be less than $\frac{W_{Lane}}{2vcos(\alpha_{max})}$, where $W_{Lane}$ is the lane width, $v$ is the vehicle speed and $\alpha_{max}$ denotes the maximum turning angle of the vehicle. 
\end{prop}

By considering the vehicle's longitudinal speed $v$ and the angle of the vehicle heading $\alpha_{max}$, the consecutive warning intervals for lane departure should be shorter than the time for the vehicle center crossing the edge of the line. For human-driven vehicles, a stricter maximum limit might be required due to possible zig-zag trajectories from unstable control.

\begin{algorithm}
\KwIn{$Z_{left}$, $Z_{right}$, pos$_{prev}$, LIKELIHOOD$\rule{0.2cm}{0.02mm}$FN()}
\KwOut{pos}
 \nl   initialize:  pos $\leftarrow$ pos$_{prev}$\;
 \While {}{
    \nl iteratively adjust `pos` to maximize the `likelihood` function\;
    %\nl gradient = gradient/(2*h)\;
    %\nl pos = pos + lr * gradient\;
    \nl \uIf{stopping$\rule{0.2cm}{0.02mm}$criteria==True}{break}
 }

 \nl \Return{pos}
 \newline
\begin{algorithmic}
\Function{likelihood$\rule{0.2cm}{0.02mm}$fn}{$Z_{left}$, $Z_{right}$, pos, $\theta$}
  \State likelihood$\rule{0.2cm}{0.02mm}$l = prob(counts$\rule{0.2cm}{0.02mm}$l, -pos, $\theta$)\;
  \State likelihood$\rule{0.2cm}{0.02mm}$r = prob(counts$\rule{0.2cm}{0.02mm}$r, pos, $\theta$)\;
  \State likelihood = likelihood$\rule{0.2cm}{0.02mm}$l * likelihood$\rule{0.2cm}{0.02mm}$r
  \State \Return likelihood
\EndFunction
\end{algorithmic}
 \caption{MLE for RF-lane positioning~\cite{suo2023rf}}
 
\end{algorithm}

With these two propositions, a detailed Algotithm 1 can be positioned which takes tag counts from both left and right antenna $Z_{left}$, $Z_{right}$, previous position pos$_{prev}$, and a likelihood function likelihood$\rule{0.2cm}{0.02mm}$fn() that depicts the relation between tag counts and vehicle positions. The output generated is the relative position of the target vehicle \textbf{pos} within the lane, and is updated at each interval based on the principle of maximum likelihood, aiming to optimize the joint likelihood of observed tag counts for enhanced accuracy.

A notable limitation of the Algorithm 1 lies in the likelihood function LIKELIHOOD$\rule{0.2cm}{0.02mm}$FN(counts$\rule{0.2cm}{0.02mm}$l, counts$\rule{0.2cm}{0.02mm}$r, pos, $\theta$). Comprehensive road testing Fig.~\ref{fig:readrate} is required to tailor to specific configurations including vehicle dimensions, antenna mount, and lane types. This is crucial for determining the empirical distribution of tag counts based on the car's lane position.

It is also worth mentioning that in practical implementation, if we have prior knowledge of tag count distribution, we can derive an analytical solution for maximum likelihood estimation and thus reduce the time complexity to linear. In particular, for a Binomial distribution of RF-lane tag counts, we can derive the vehicle's relative position estimation in constant time.

\begin{figure*}[tb!]
\centering
\begin{subfigure}{0.305\textwidth}
  \centering
  \includegraphics[width=\textwidth]{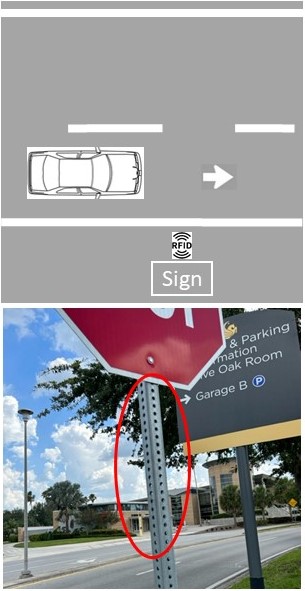}
  \caption{Scenario 1}
  \label{fig:tagnum_s1}
\end{subfigure}
\begin{subfigure}{0.305\textwidth}
  \centering
  \includegraphics[width=\textwidth]{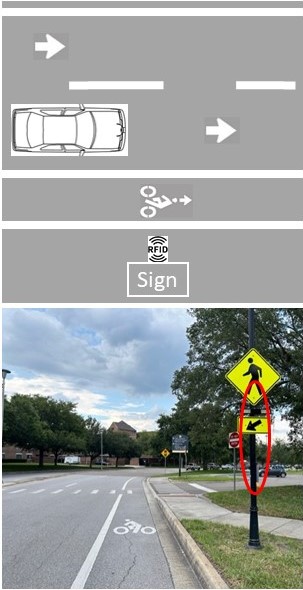}
  \caption{Scenario 2}
  \label{fig:tagnum_s2}
\end{subfigure}
\begin{subfigure}{0.305\textwidth}
  \centering
  \includegraphics[width=\textwidth]{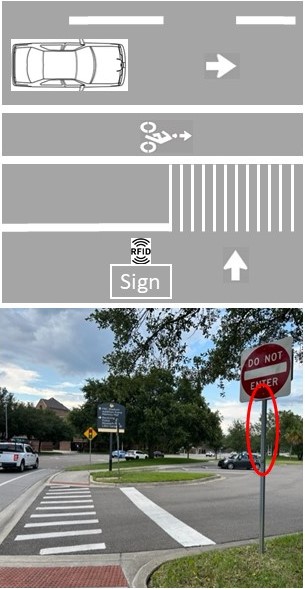}
  \caption{Scenario 3}
  \label{fig:tagnum_s3}
\end{subfigure}
\begin{subfigure}{0.305\textwidth}
  \centering
  \includegraphics[width=\textwidth]{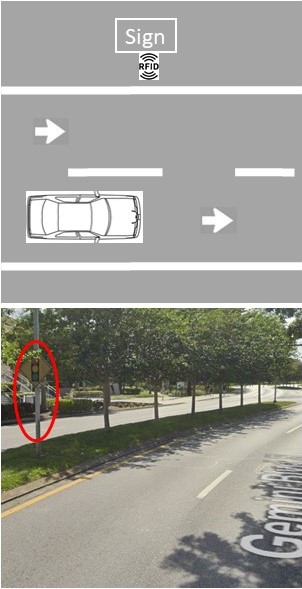}
  \caption{Scenario 4}
  \label{fig:tagnum_s4}
\end{subfigure}
\begin{subfigure}{0.305\textwidth}
  \centering
  \includegraphics[width=\textwidth]{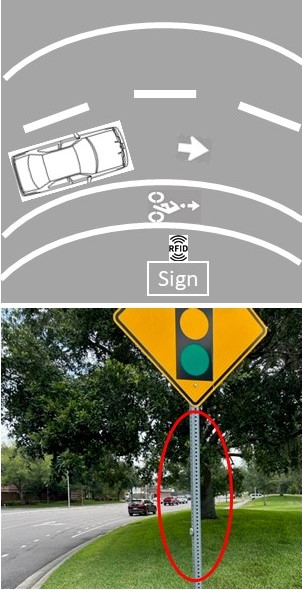}
  \caption{Scenario 5}
  \label{fig:tagnum_s5}
\end{subfigure}
\begin{subfigure}{0.305\textwidth}
  \centering
  \includegraphics[width=\textwidth]{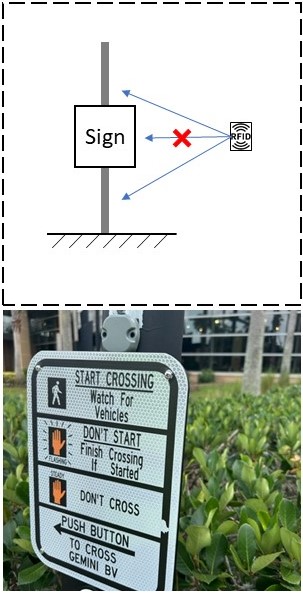}
  \caption{Scenario 6}
  \label{fig:tagnum_s6}
\end{subfigure}
\caption{Traffic scenarios for vehicle-sign communications under different road geometries and sign placement}
\label{fig:trafficsign_scenario}
\end{figure*}

\begin{figure*}[tb!]
\centering
\includegraphics[width=0.8\textwidth]{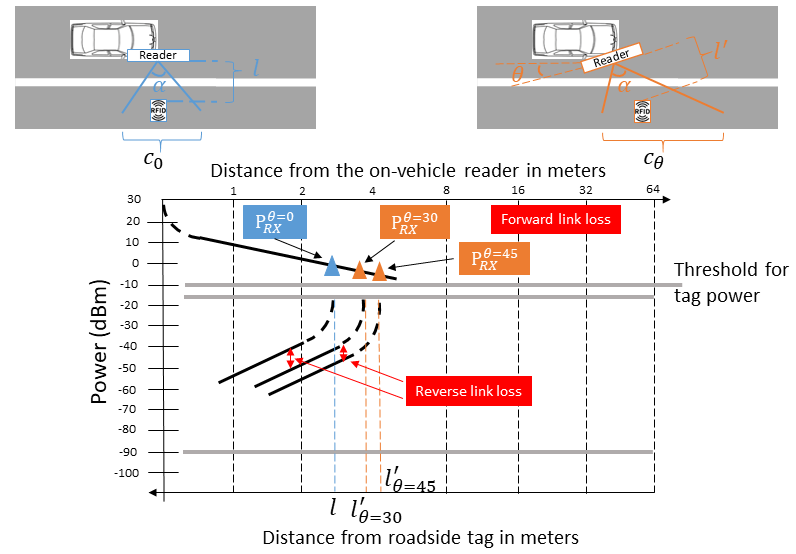}
\caption{The tradeoff between reader coverage and path loss}
\label{fig:tradeoff}
\end{figure*}

\subsection{Traffic sign inventory management}
\paragraph{Motivation}
Traffic sign inventory management calls for efficient detection and recognition of different signs. While there exist studies on automatic sign recognition, these efforts often only consider a small subset of safety-related traffic signs~\cite{stallkamp2011german}, or necessitate a large training dataset for deep learning models to automate sign classification~\cite{tabernik2019deep}. 

We propose a more efficient and extensible approach for traffic sign detection for inventory management—encoding traffic rules within RFID tags. These tags can then be interrogated by passing vehicles, which can transmit related information to a cloud server for real-time inventory updates~\cite{garcia2018passive,chen2022rfid}.

\paragraph{Preliminary designs and challenges}\label{section:sign_challenge}

Unlike traffic sign detection for autonomous driving, sign detection for inventory management needs to accommodate various placements of signs across heterogeneous road geometries, as depicted in Fig.~\ref{fig:trafficsign_scenario}. The reason is that determining the link budget of REI needs to account for different vehicle-sign distances under heterogeneous road geometries and the uncertainties in the installation height of signs. In particular, we evaluate REI designs through real-world experiments to answer three research questions:

\begin{itemize}
    \item \textbf{Q1}: To what extent will the differences in vehicle-sign distances affect the reading performance?
    \item \textbf{Q2}: How will different installation heights of traffic signs influence the reading reliability?
    \item \textbf{Q3}: How should the on-vehicle antenna be designed to optimize the reading performance? 
\end{itemize}

The tradeoff discussed above has important implications for designing the link budget of the system. Ideally, the antenna orientation angle, $\theta$, should be maximized to ensure that the received power, $P_{RX}$, is greater than the minimum threshold needed to power the tag's integrated circuit, as depicted in Fig.~\ref{fig:tradeoff}. Furthermore, given that the RF signal's strength diminishes in proportion to the fourth power of the distance between the tag and reader, the on-vehicle reader must possess sufficient sensitivity to detect the backscattered signal from the tag.

In answering Q1, we consider five typical scenarios, each of which has a unique road geometry as shown in  Fig.~\ref{fig:tagnum_s1}-\ref{fig:tagnum_s4}. Scenarios 1-4 involve different vehicle-sign distances, resulting in different path losses for backscatter signals. A curved road is shown in Fig.~\ref{fig:tagnum_s5}: the greater the curvature of the road, the more time it takes for the tag to leave the coverage area of the on-vehicle reader, increasing the potential of successful reads.

Additionally, the tag installation height might not always be possible, as shown in Fig.~\ref{fig:tagnum_s6}. While positioning on-road tags at the same height as vehicle-mounted readers can optimize power strength for data transmission, it may not be permissible to attach RFID tags on certain pedestrian crosswalk signs that are lower than other signs due to safety considerations, as such installations could potentially obstruct the view of drivers or pedestrians."

Addressing Q3 requires balancing reader coverage with path loss, as illustrated in Fig.~\ref{fig:tradeoff}. Aligning the reader antenna perpendicular to the vehicle’s longitudinal direction can enhance the RF signal's power received by tags along the road. However, this alignment reduces the reader's effective coverage area. Section V-B-b will demonstrate that the optimal orientation for the on-vehicle antenna lies between the vehicle's longitudinal and lateral directions to optimize reading time, with path loss being a secondary consideration. 

The enhancement in the effective reading area as we expand the mounting angle can be derived as follows. Consider the reader's mounting angle to be zero when it is perpendicular to the longitudinal direction, as depicted in the upper-left of Fig.~\ref{fig:tradeoff}. Assuming the distance between the reader and the tag to be $l$ and the antenna angle to be $\alpha$, and that an isotropic antenna is in use, we can calculate the effective length of the reader's coverage area in the longitudinal direction as follows.

\begin{equation}\label{eq:readlength_zero}
    c_0=2ltan\frac{\alpha}{2}
\end{equation}

As we increase and set the mounting angle to be $\theta$, as shown on the upper-right of the right of Fig.~\ref{fig:tradeoff}, the effective length of the reader coverage area will also become larger, as given in eq.~\ref{eq:readlength_theta}. For instance, in a vehicle configuration where the antenna's beamwidth, $\alpha$, is 60 degrees and the distance between the vehicle and the traffic sign is three meters, the length of the reader coverage area is approximately 3.464 meters when $\theta$ equals zero. However, when $\theta$ is increased to 30 degrees, the effective length expands to 5.196 meters.

\begin{equation}\label{eq:readlength_theta}
    C_{\theta} = ltan(\frac{\alpha}{2}-\theta) + ltan((\frac{\alpha}{2}-\theta)
\end{equation}

Although enlarging the mounting angle increases the effective length and, therefore, the time available for the on-vehicle reader to pick up each tag, it also lengthens the distance between the antenna and tags. This lengthening reduces the power strength of the RF signal available for energy harvesting and backscattering. Assuming that the original antenna-tag distance is $l$, the enlarged distance $l'$ can be calculated as $\frac{l}{cos(\theta)}$. Therefore, we can derive the difference in path loss between the original (zero) and the enlarged antenna angle ($\theta$) as follows.

\begin{equation}\label{eq:enlarged_dist}
    \Delta P = 40log_{10}\frac{l}{cos(\theta)} - 40log_{10}l
\end{equation}

\subsection{Crowdsensing for road weather conditions}

\begin{figure*}[tb!]
\centering
\begin{subfigure}{0.35\textwidth}
  \centering
  \includegraphics[width=\textwidth]{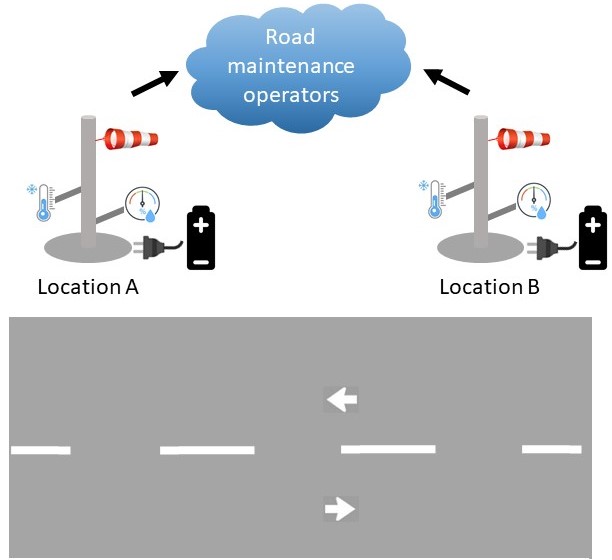}
  \caption{Client-server based on fixed-location ESS}
  \label{fig:cs_arch}
\end{subfigure}
\begin{subfigure}{0.4\textwidth}
  \centering
  \includegraphics[width=\textwidth]{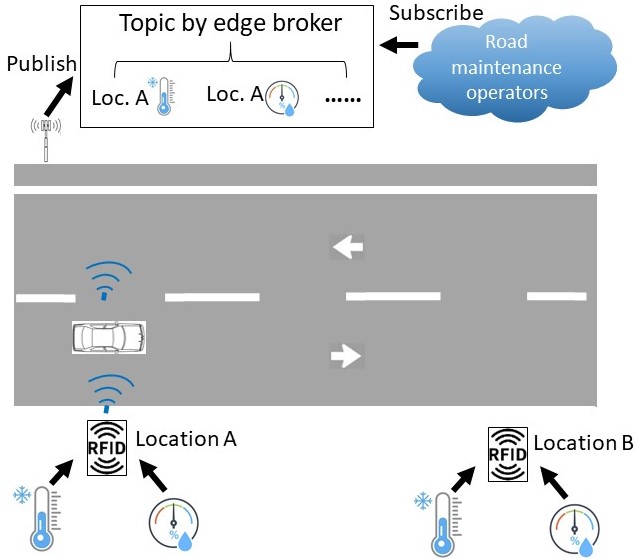}
  \caption{pub-sub based on on-road RFID and vehicle sensors}
  \label{fig:ps_arch}
\end{subfigure}
\caption{REI-enabled crowdsensing architecture for road weather condition monitoring}
\label{fig:sensing_arch}
\end{figure*}

Sensing and disseminating information about weather conditions on or near road surfaces is crucial to traffic safety and efficiency. This is because real-time weather information can be used to alert drivers of potential road hazards~\cite{Alfelorweather2005} or influence their decisions on departure time and route selection~\cite{FHWAweather}. In the U.S., Road Weather Information Systems (RWIS) are proposed to collect and disseminate sensing data collected from field sensors to provide decision support to road operators.

Previously, the RWIS were based on a client-server architecture where a cloud-based road weather information system leveraged data collected through Environmental Sensor Stations (ESS) to track and make predictions on the changes in road and weather conditions, as shown in Fig.~\ref{fig:cs_arch}. The information will be used by highway operators and maintenance staff to assess the current and future impact on highways and respond to hazardous conditions. One example is the road maintenance managers who need to decide on anti-icing materials and treatment strategies after abrupt changes in temperature and humidity are detected on pavement surfaces in a given road segment~\cite{manfredi2008road}. Although ESS provides accurate weather information for the forecast of pavement surface conditions, their placement locations are constrained by the deployment cost and ease of access to power and communication infrastructure for transmitting sensing data. In fact, ESS is often installed at fixed locations based on the experience of field operators~\cite{jin2014determining}.

Compared to the RWIS based on a client-server architecture discussed earlier, the proposed REI provides updates on road weather conditions with lower costs and an energy-efficient manner, as shown in Fig.~\ref{fig:ps_arch}. Without power supplies from batteries, the crowdsource sensing solution assisted by REI harvests energy from the interrogation signals from on-vehicle readers. For piezoelectric sensors that are embedded within infrastructure, previous work shows that they are capable of energy scavenging from strain-variations~\cite{aono2016infrastructural}. Furthermore, for RFID sensors that are deployed on the roadside, the microcontroller may even harvest energy from the interrogation signal received by the transponder~\cite{costa2021review}.

In addition to energy efficiency, the REI-enabled crowdsensing RWIS is more flexible in changing the message routes between sensing data producers (i.e., vehicles) and consumers (i.e., stakeholders in weather information). We recommend the use of a publish-subscribe pattern to realize RWIS as shown in Fig.~\ref{fig:ps_arch}. The architecture is more flexible than the client-server architecture in that a “data publisher” (i.e., a vehicle) can only specify the topic it is going to publish without deciding on the exact receiver to which it will send messages. A sensing data receiver, on the other hand, may only subscribe to an event (road weather conditions in a given location) that is interesting to it without knowing who actually generates such information. This reduces the burden of adding new topics of interest.

\begin{table}[t]
	\caption{COTS RFID equipment used in the experiment evaluation}
		\begin{tabularx}{\columnwidth}{c|X}   
			\hline
			\textbf{Hardware component}&\textbf{Description} \\
			\hline
			\hline
			RFID Reader&Impinj R700  \\
			\hline
			Reader antenna&Laird S9025PR outdoor antenna \\
                \hline
			  Sessions&Session 0\\
                \hline
                Search Mode&Dual target\\
			\hline
			Tag 1&HID Epoxy UHF RFID Tag\\
                \hline
			Tag 2&Omni-ID Exo 750 RFID Tag\\
			\hline
                \hline
                \textbf{Reader settings}&\textbf{Description} \\
                \hline
			Power level&30 dBm \\
			\hline
			Receiver sensitivity&90 dBm \\
                \hline
                Data encoding schemes&FM0,Miller 2,4,8\\
			\hline

		 \end{tabularx}
		\label{tab:hardware}
\end{table}

\section{Experimental evaluation}

\begin{figure*}[tb!]
\centering
\begin{subfigure}{0.35\textwidth}
  \centering
  \includegraphics[width=\textwidth]{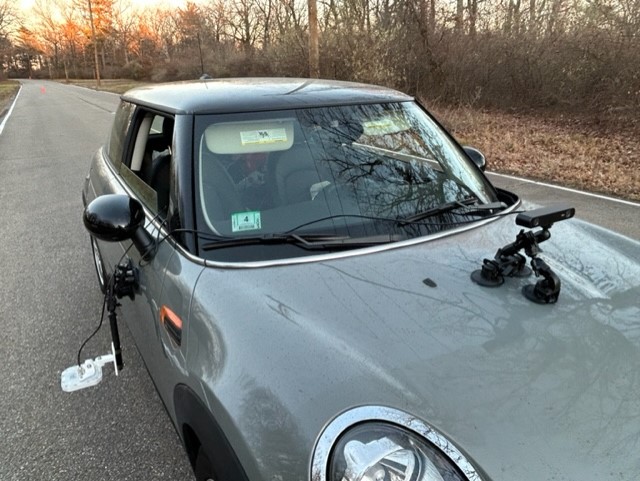}
  \caption{REI for lane-markings experiments}
  \label{fig:car_mount1}
\end{subfigure}
\begin{subfigure}{0.35\textwidth}
  \centering
  \includegraphics[width=\textwidth]{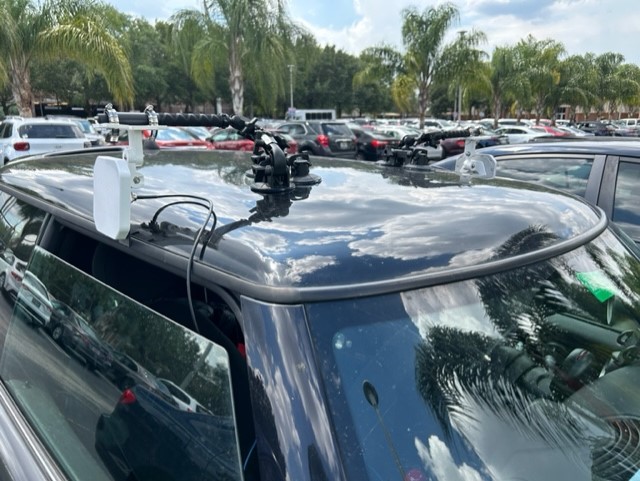}
  \caption{REI for traffic sign inventory management}
  \label{fig:car_mount2}
\end{subfigure}
\caption{The placement of on-vehicle antennas}
\label{fig:car_mounts}
\end{figure*}

\begin{figure*}[tb!]
\centering
\begin{subfigure}{0.3\textwidth}
  \centering
  
  \includegraphics[width=\textwidth]{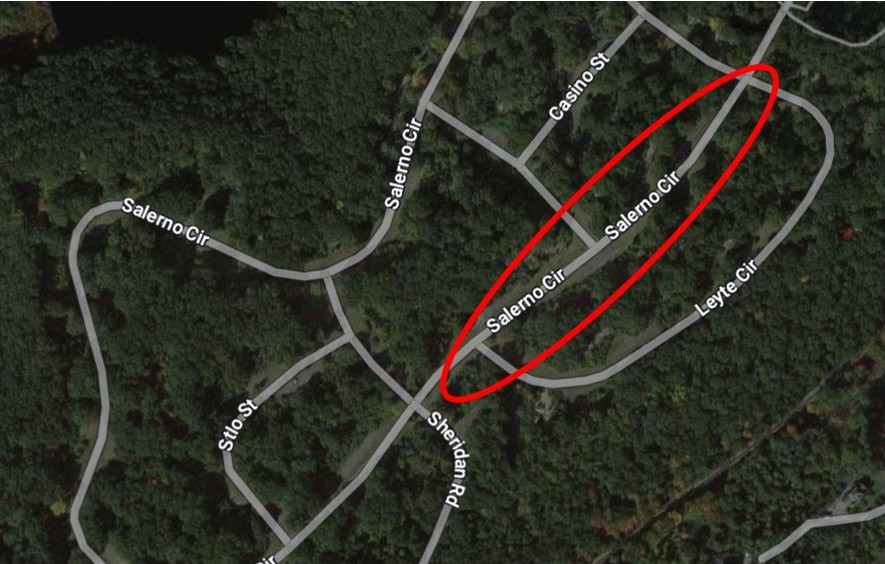}
  \caption{The closed track in MA, USA}
  \label{fig:testing_road1}
\end{subfigure}
\begin{subfigure}{0.3\textwidth}
  \centering
  \includegraphics[width=\textwidth]{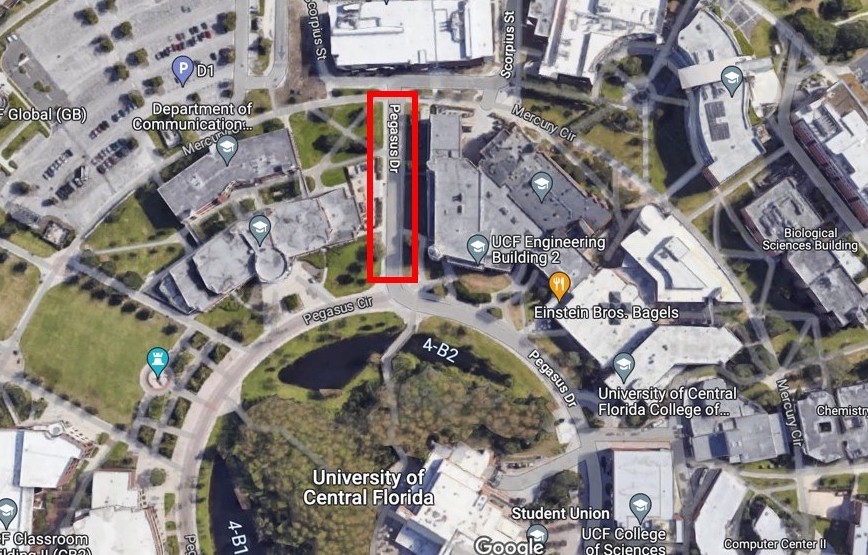}
  \caption{The straight-line road within UCF}
  \label{fig:testing_road2}
\end{subfigure}
\begin{subfigure}{0.3\textwidth}
  \centering
  \includegraphics[width=\textwidth]{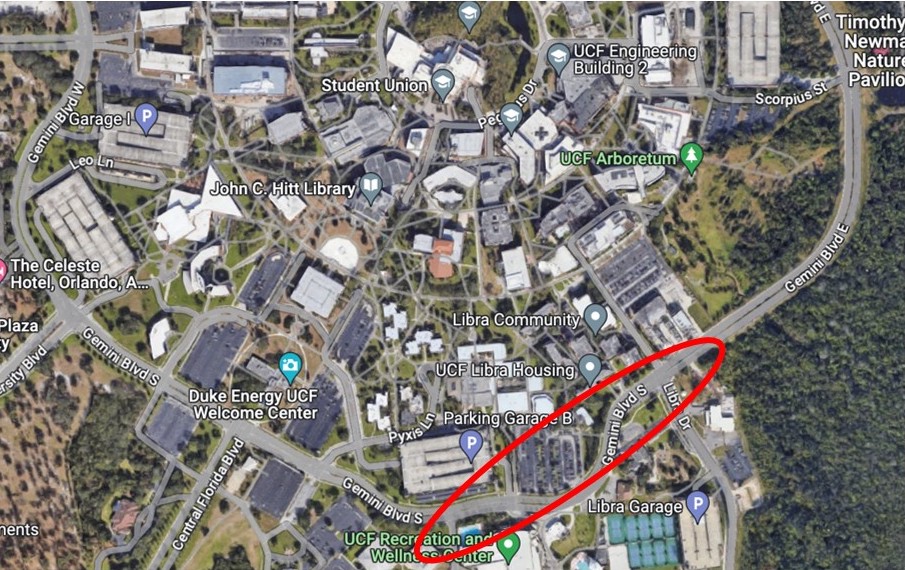}
  \caption{The ring loop road around UCF}
  \label{fig:testing_road3}
\end{subfigure}
\caption{Closed and public roads for experimental evaluations}
\label{fig:testing_roads}
\end{figure*}

\subsection{Experimental settings}
The configurations of the testing vehicle are shown in Fig.~\ref{fig:car_mounts}. In the former case, the antennas are attached to each side of the vehicle while facing downward to capture the tags deployed on the road surface as lane markings, as shown in Fig.~\ref{fig:car_mount1}. The second configuration involves placing antennas such that their signal propagation direction runs parallel to the earth's surface, as shown in Fig.~\ref{fig:car_mount2}. We also have an RFID reader and a computing platform within the testing vehicle for real-time decoding and processing of data backscattered from on-road tags. The RFID readers, tags and antennas we used in our experiments are given in Table~\ref{tab:hardware}.

For the experimental evaluation of REI, we selected a closed testing track and two public roads. The closed testing track, known as Salerno Circle, is located in Western Massachusetts, USA, and is part of the \textit{Devens UAV} test site for autonomous vehicle operation (see Fig.~\ref{fig:testing_road1}). 

The public roads, on the other hand, are situated near the campus of the University of Central Florida (UCF) in Orlando, FL, USA. The first one, as shown in Fig.~\ref{fig:testing_road2}, was used to mimic the influence of background objects on lane-marking experiments. For example, the multi-path effect incurred by pedestrians, surrounding vehicles, trees, and buildings might create noises on read-tag communication, degradating REI performance. The second public road is part of a loop road around the UCF campus (Fig.~\ref{fig:testing_road3}). This loop road features varying geometries concerning sign placement, enabling us to test vehicle-sign communication and determine the optimal configuration for reading performance.

\begin{figure*}[tb!]
\centering
\includegraphics[width=0.60\textwidth]{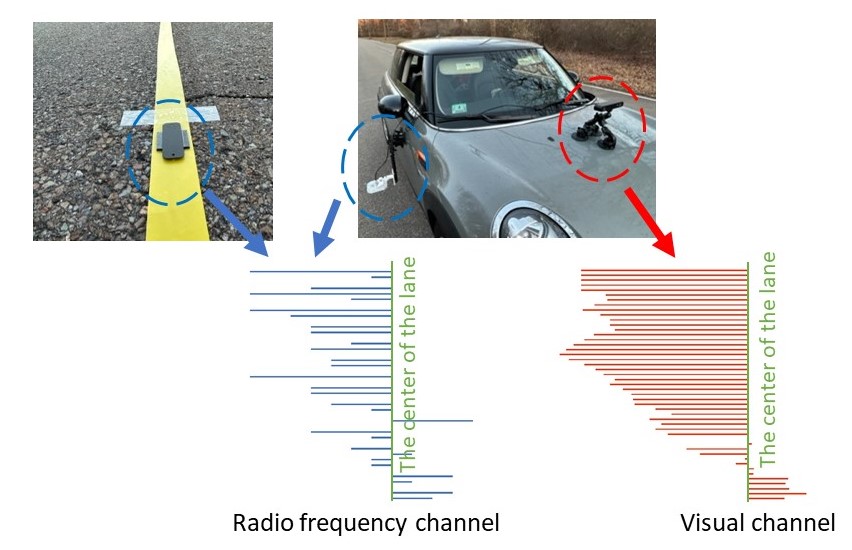}
\caption{The experimental setup of the RF pavement marking system and example results~\cite{suo2023rf}}
\label{fig:experiment_setup}
\end{figure*}

\begin{figure*}[tb!]
\centering
\begin{subfigure}{0.28\linewidth}
  \centering
  \includegraphics[width=\linewidth]{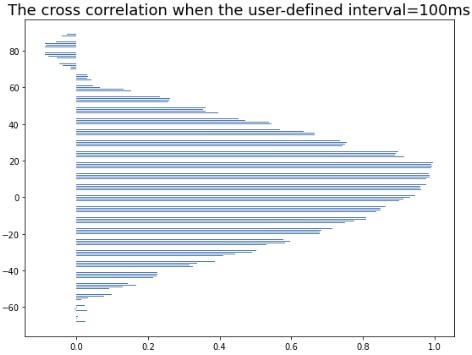}
  \caption{Vehicle moving speed = 10mph}
  \label{fig:corr_10mph}
\end{subfigure}
\begin{subfigure}{0.28\linewidth}
  \centering
  \includegraphics[width=\linewidth]{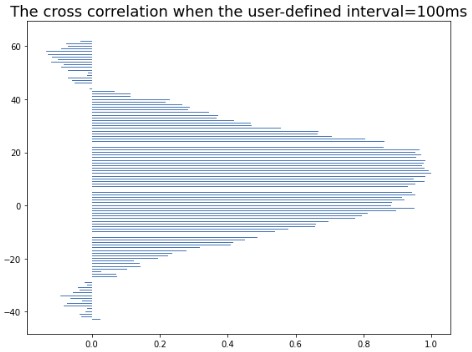}
  \caption{Vehicle moving speed = 20mph}`
  \label{fig:corr_20mph}
\end{subfigure}
\begin{subfigure}{0.28\linewidth}
  \centering
  \includegraphics[width=\linewidth]{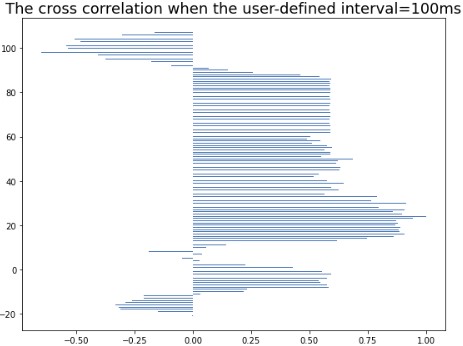}
  \caption{Vehicle moving speed = 40mph}
  \label{fig:corr_40mph}
\end{subfigure}
\caption{The normalized cross-correlation of lane detection results between the RF and vision-based systems~\cite{suo2023rf}.} 
\label{fig:correlation}
\end{figure*}

\begin{figure*}[tb!]
\centering
\begin{subfigure}{0.32\textwidth}
  \centering
  \includegraphics[width=\textwidth]{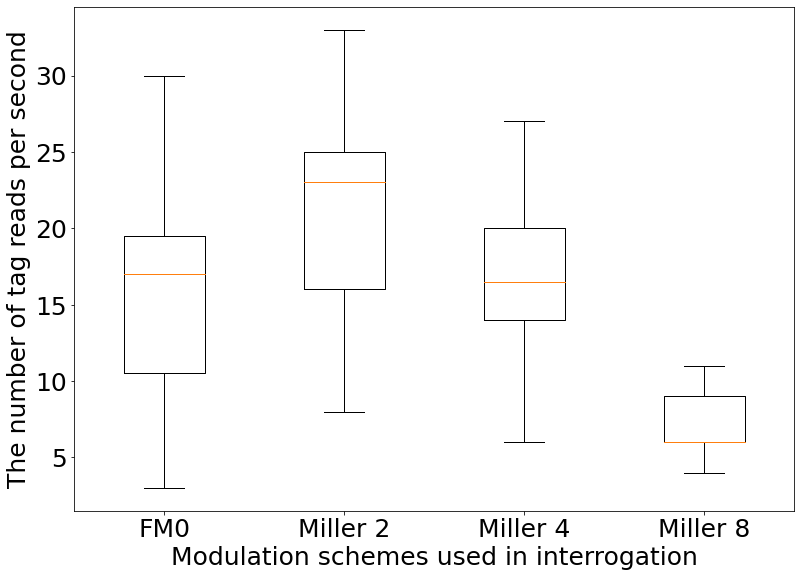}
  \caption{Vehicle speed=10mph}
  \label{fig:lane_modulation_10mph}
\end{subfigure}
\begin{subfigure}{0.32\textwidth}
  \centering
  \includegraphics[width=\textwidth]{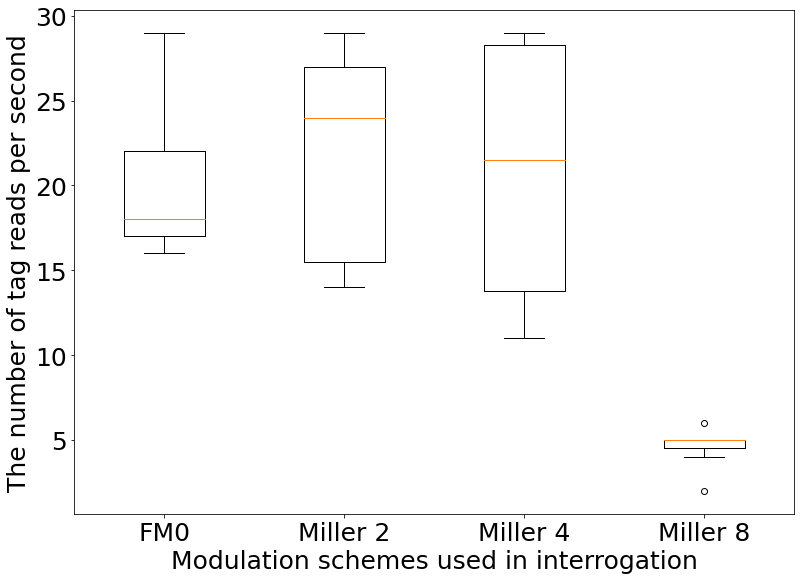}
  \caption{Vehicle speed=20mph}
  \label{fig:lane_modulation_20mph}
\end{subfigure}
\begin{subfigure}{0.32\textwidth}
  \centering
  \includegraphics[width=\textwidth]{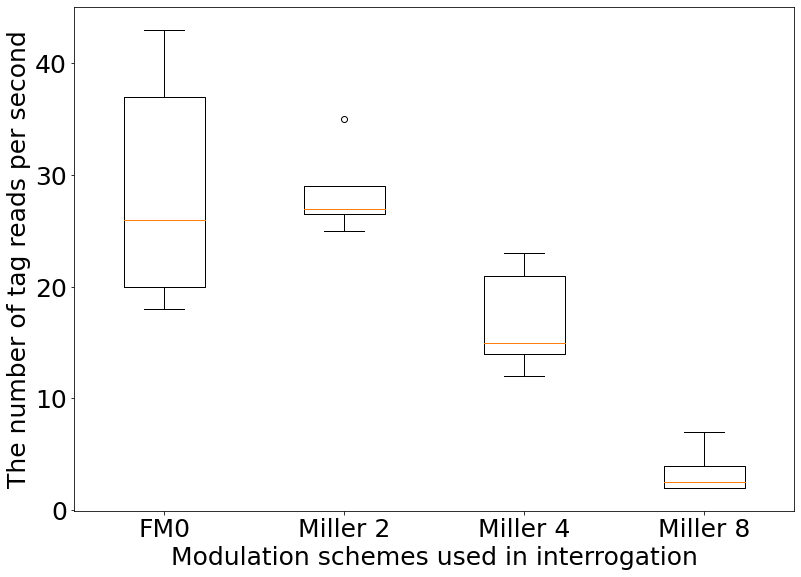}
  \caption{Vehicle speed=35mph}
  \label{fig:lane_modulation_35mph}
\end{subfigure}
\caption{Performance variations among different modulation schemes during lane tag interrogation}
\label{fig:encoding_scheme}
\end{figure*}

\subsection{Results and discussion}

\paragraph{The influence of encoding scheme}
We rely on testing results from the lane marking experiments to evaluate the influence of vehicle speed on REI performance. Compared to other REI applications, such as traffic sign inventory management or weather-condition sensing, the RF lane marking is more sensitive to changes in vehicle speed. The reason is that the proposed filtering algorithm (Algorithm 1) leverages repetitive readings from on-road tags to infer the vehicle's relative position to the edges of the lane. Therefore, missing readings can result in erroneous predictions. 

As depicted in Fig.~\ref{fig:encoding_scheme}, the Miller 2 encoding scheme outperforms other Miller encoding schemes, even those with more clock cycles during which no signal state change may occur. Surprisingly, the Miller 2 scheme is even more beneficial than the FM0 scheme, which was anticipated to produce a higher throughput regarding the number of RFID reads. Our interpretation of these results is that the Miller 2 scheme is better at balancing throughput and noise cancellation, which is crucial for moving vehicles operating in outdoor environments.

\paragraph{The influence of vehicle speed and reader coverage}
The performance of REI significantly varies across different antenna placements and vehicle speed. 

We began our experiments with an RF lane-marking system to evaluate the impact of vehicle speed, as shown in Fig.~\ref{fig:experiment_setup}. In addition to the RF link between the vehicle reader and the tags deployed on the surface of the roads, we also installed a ZED 2 camera on the front of the vehicle to unique lane detection results from the visual channel as the benchmark. The vertical lines in Fig.~\ref{fig:experiment_setup} represent the center of the lane, while the horizontal bars indicate the testing vehicle's relative distance from the center.

Our lane-marking experiments revealed a notable degradation in REI performance at vehicle speeds exceeding 40 mph when antennas were positioned facing downward. Figure~\ref{fig:correlation} depicts the normalized cross-correlation between RF-based and vision-based lane detection systems. Figures\ref{fig:corr_10mph} and~\ref{fig:corr_20mph} demonstrate a close alignment between RF and vision outputs at speeds of 10 and 20 mph. However, Fig.~\ref{fig:corr_40mph} illustrates a mismatch and, thus, a performance decline when the vehicle's speed is increased to 40 mph.

\begin{figure*}[tb!]
\centering
\begin{subfigure}{0.32\textwidth}
  \centering
  \includegraphics[width=\textwidth]{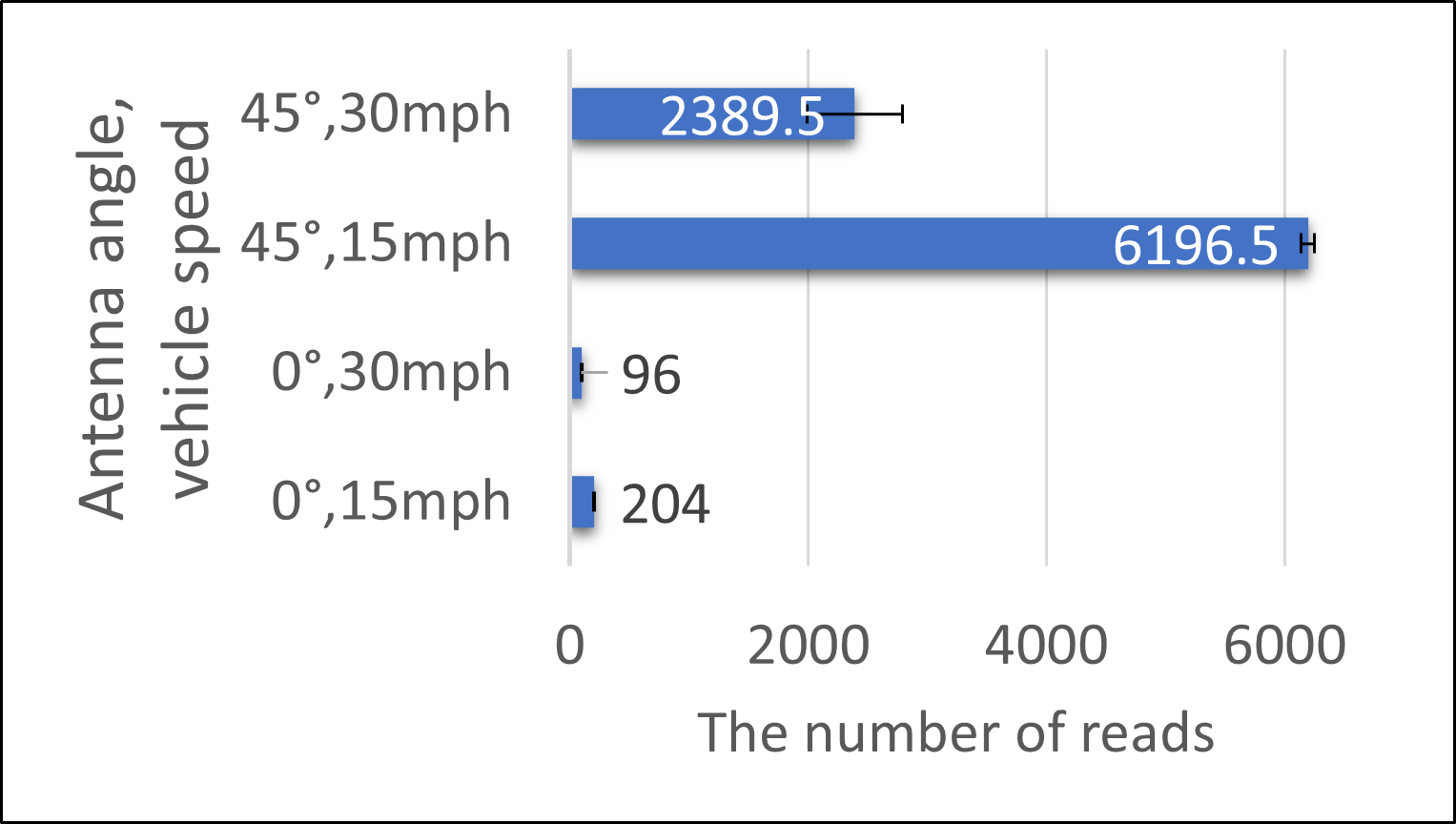}
  \caption{Scenario 1}
  \label{fig:trafficsign_num_s1}
\end{subfigure}
\begin{subfigure}{0.32\textwidth}
  \centering
  \includegraphics[width=\textwidth]{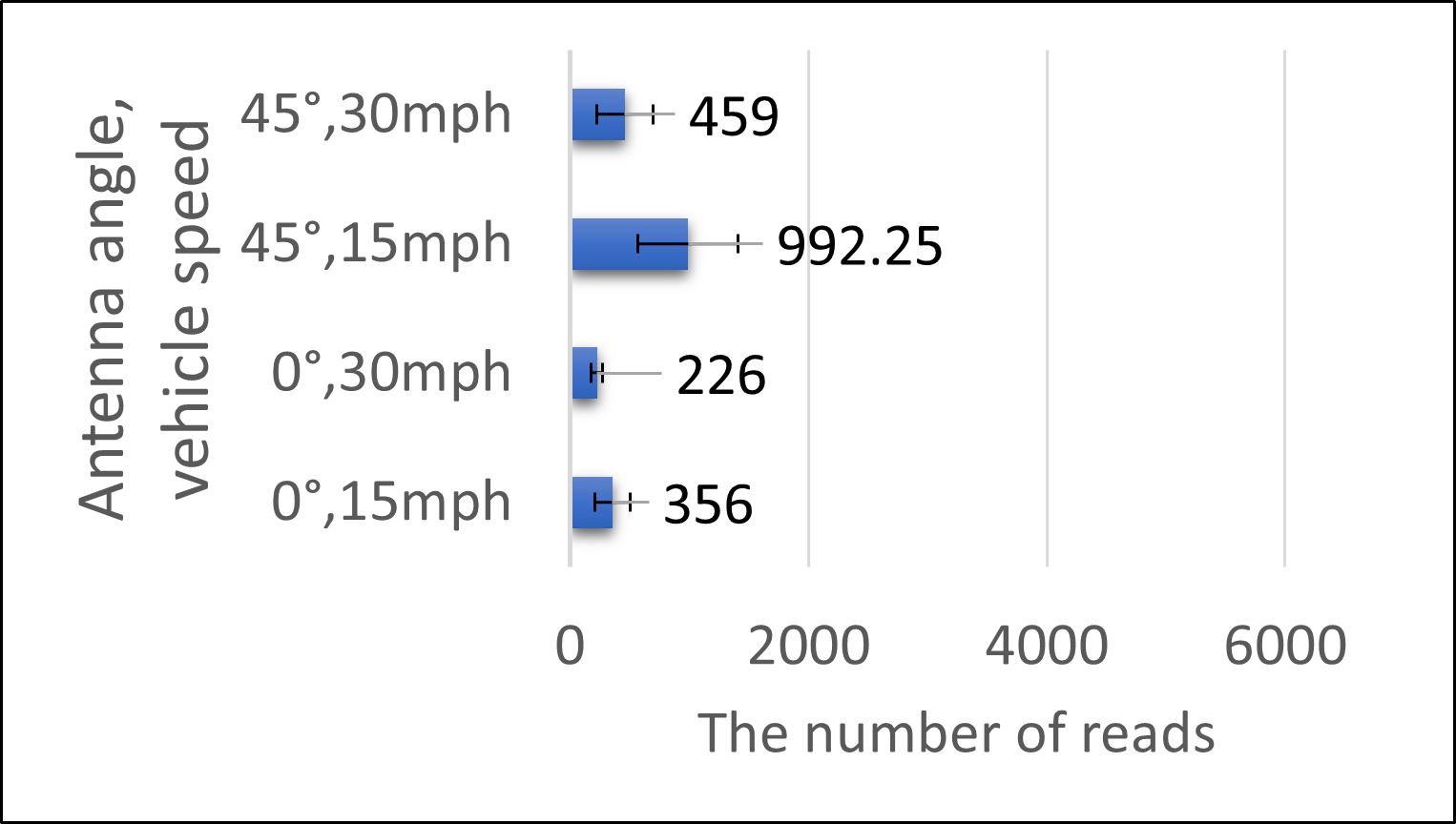}
  \caption{Scenario 2}
  \label{fig:trafficsign_num_s2}
\end{subfigure}
\begin{subfigure}{0.32\textwidth}
  \centering
  \includegraphics[width=\textwidth]{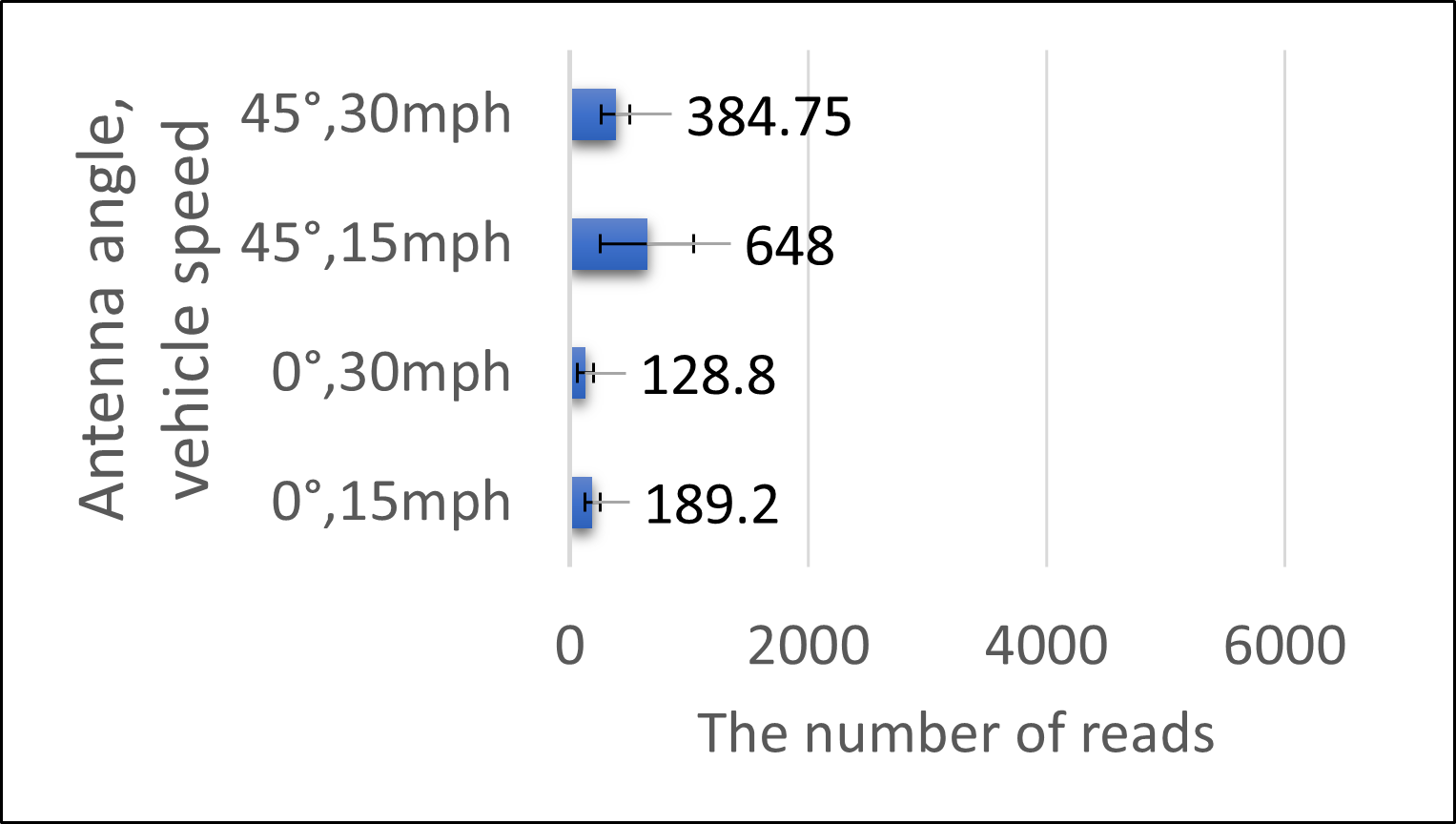}
  \caption{Scenario 3}
  \label{fig:trafficsign_num_s3}
\end{subfigure}
\begin{subfigure}{0.32\textwidth}
  \centering
  \includegraphics[width=\textwidth]{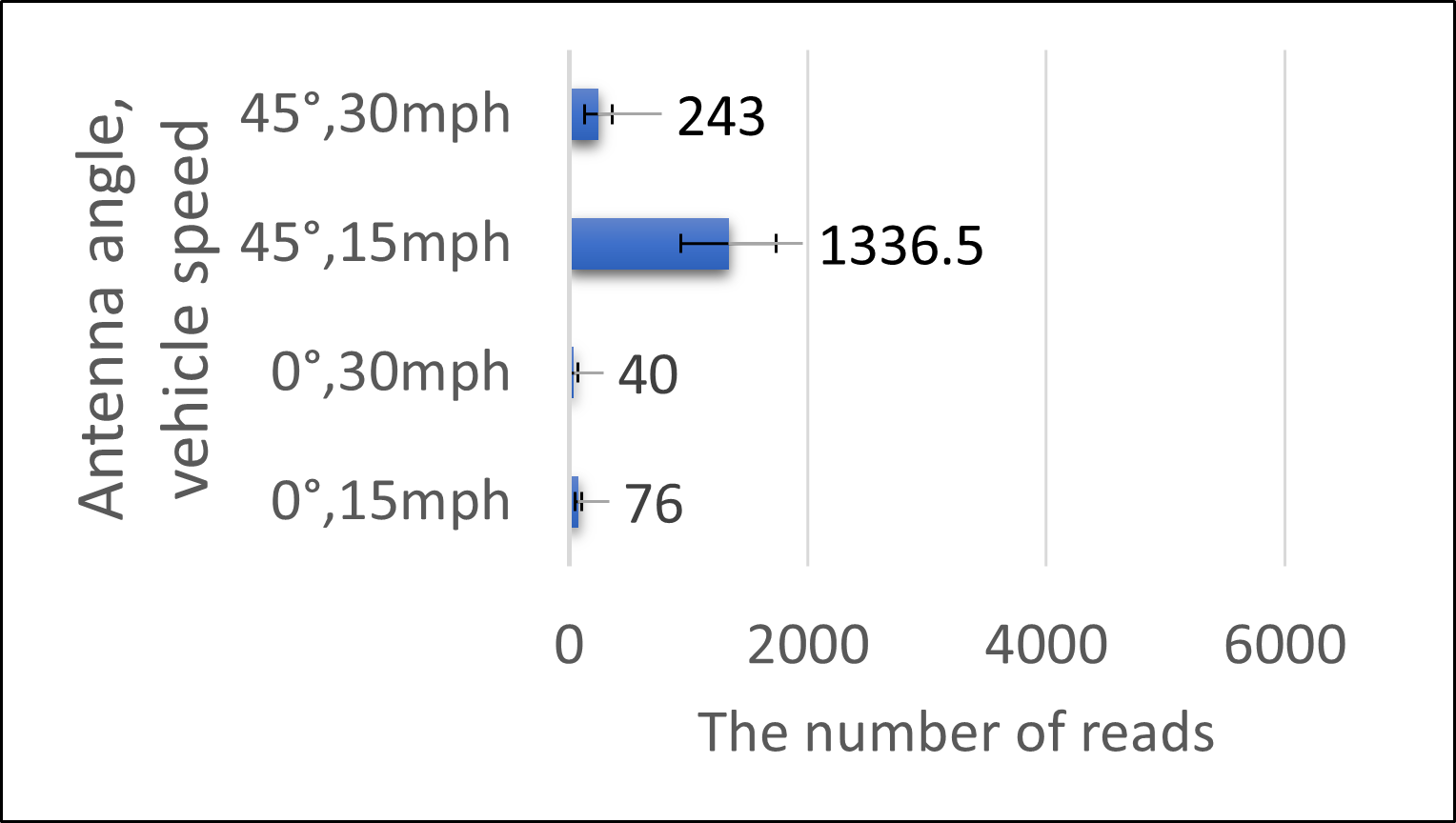}
  \caption{Scenario 4}
  \label{fig:trafficsign_num_s4}
\end{subfigure}
\begin{subfigure}{0.32\textwidth}
  \centering
  \includegraphics[width=\textwidth]{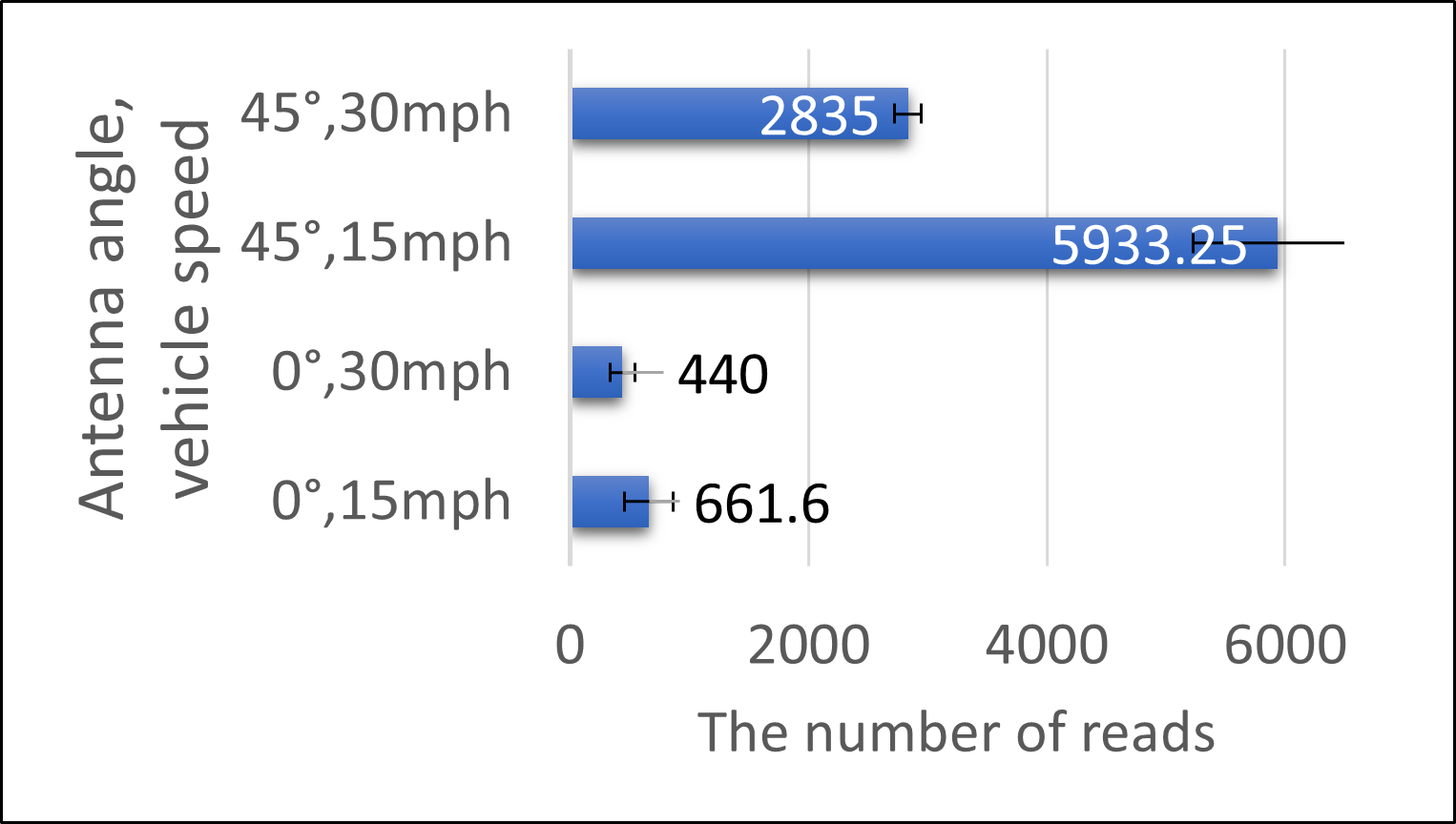}
  \caption{Scenario 5}
  \label{fig:trafficsign_num_s5}
\end{subfigure}
\begin{subfigure}{0.32\textwidth}
  \centering
  \includegraphics[width=\textwidth]{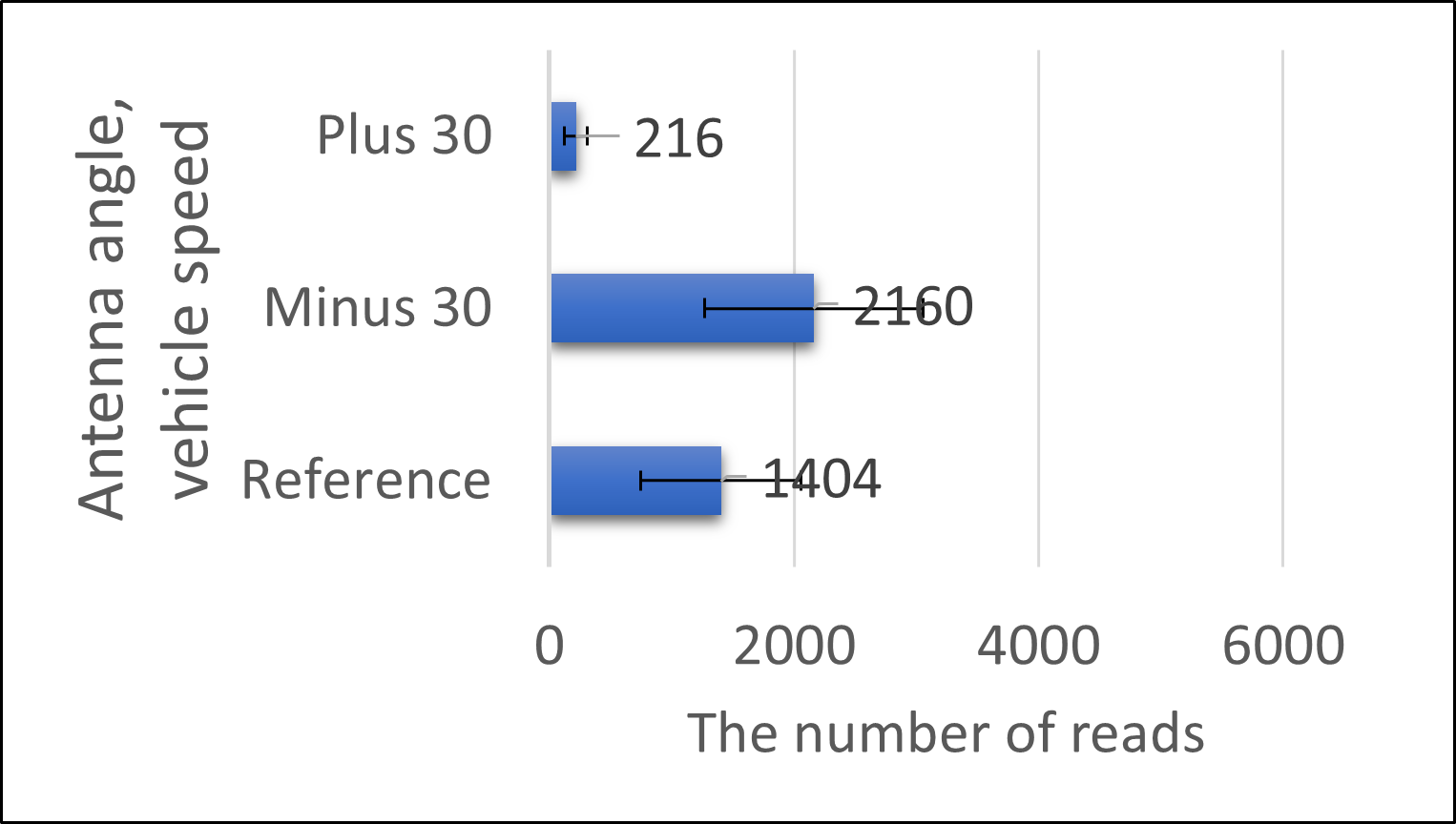}
  \caption{Scenario 6}
  \label{fig:trafficsign_num_s6}
\end{subfigure}
\caption{The number of traffic sign readings under different antenna angles and vehicle speeds}
\label{fig:trafficsign_num}
\end{figure*}

Similarly, in traffic sign inventory management scenarios, where the antenna signal propagation direction is parallel to the earth's surface, a higher vehicle speed also leads to fewer tag reads, as shown in Fig.~\ref{fig:trafficsign_num}. For example, in the traffic-sign scenario 1 depicted in Fig.~\ref{fig:tagnum_s1}, the number of reads, when the vehicle moves at 15 mph, is twice as much as 30 mph for all antenna configurations, as shown in Fig.~\ref{fig:trafficsign_num_s1}. Similar trends are observed in scenarios 2-5, as shown in Fig.~\ref{fig:trafficsign_num_s1}-\ref{fig:trafficsign_num_s5}.

\paragraph{The influence of path loss by road geometries}
In answering Q1 presented in Section IV-B-b, i.e., the influence of path loss induced by road geometries on REI performance, we compare the number of readings across different scenarios, as shown in Fig.~\ref{fig:trafficsign_num}. With the shortest tag-reader distance, the number of readings by the on-vehicle antenna under scenario 1 is significantly larger than in scenarios 2-4. The phenomenon can be interpreted by the fact that RF energy is dissipated at a rate proportional to the fourth power of the tag-reader distance, as illustrated in Fig.~\ref{fig:tradeoff}.

Interestingly, scenarios 1 and 5 display similar REI performance in terms of the number of tag readings. This is despite the tag-reader distance in scenario 5 being almost equivalent to that in scenario 2. We attribute this to the extended reading time afforded by the road curvature in scenario 5. This additional reading time allows the on-vehicle reader to effectively interrogate the tag in line-of-sight, taking advantage of the increased duration before the tag moves out of its coverage area.

\paragraph{The influence of the tag installation height}
The experiments conducted on traffic sign inventory management provide insights to address Question 2, as stated earlier. Specifically, road tags positioned at different heights exhibit varying degrees of susceptibility to the multi-path effect. In our tests involving three distinct heights for RFID tag installation, we maintained a vehicle speed of 15 mph and set the on-vehicle antenna angle to 45 degrees to optimize tag readings. This allows us to evaluate the influence on system performance by installation height and the multi-path effect. When the tag is placed at a lower position on the traffic sign, the primary link between the reader and the tag might encounter heightened environmental interference from RF signals reflecting off the ground, as shown in Fig.~\ref{fig:tagnum_s6}.

Interestingly, this indirect communication path could bolster the connection between the reader and the tag. This is demonstrated in Fig.~\ref{fig:trafficsign_num_s6}. When the tag is affixed to a pedestrian sign 30 centimeters below the reference height, the reader records 2,160 readings. This signifies a boost of 54 percent increases in readings compared to the reference height and a remarkable increase of 900 percent in readings when contrasted with the readings at a height 30 centimeters above the reference. This suggests that the multi-path effect could potentially be harnessed to improve reader-tag communication, especially in scenarios where tags are attached at lower positions.

\paragraph{The implication for on-vehicle antenna design}
Although vehicle speed can affect REI performance, as discussed in Section-V-B-b, its impact is substantially less significant than that of the orientation angle of the on-vehicle antenna. For example, in scenario 1, when the on-vehicle antenna is angled at 45 degrees, the number of reads is approximately 2,389 and 6,196 at speeds of 30 and 15 mph, respectively, as shown in Fig.~\ref{fig:trafficsign_num_s1}. In contrast, the on-vehicle reader only gets 96 and 204 reads with 30 and 15 mph vehicle speeds when the antenna was installed at a zero-degree angle. The same tendencies are discernible in scenarios 2-5, as shown in Fig.~\ref{fig:trafficsign_num_s1}-\ref{fig:trafficsign_num_s5}.

These results provide valuable insights into the design tradeoff between reader coverage and path loss in answering Q3, as discussed in Section-B-b and depicted in Fig.~\ref{fig:tradeoff}. We can mitigate concerns about the path loss associated with increasing the on-vehicle antenna's orientation angle by appropriately adjusting antenna power and sensitivity. This adjustment enables us to position the on-vehicle reader antenna at an angle that maximizes reader coverage, thus enhancing the probability of moving vehicles picking up roadside tags.

\paragraph{The influence of interrogation on user memory bank}
Compared to the first two use cases, REI has a longer inventory cycle when it is used for sensing road weather conditions. The reason lies in the extra time incurred by reading RFID sensor data from user memory bank within each tag.

\begin{figure}[tb!]
\centerline{\includegraphics[width=0.48\textwidth]{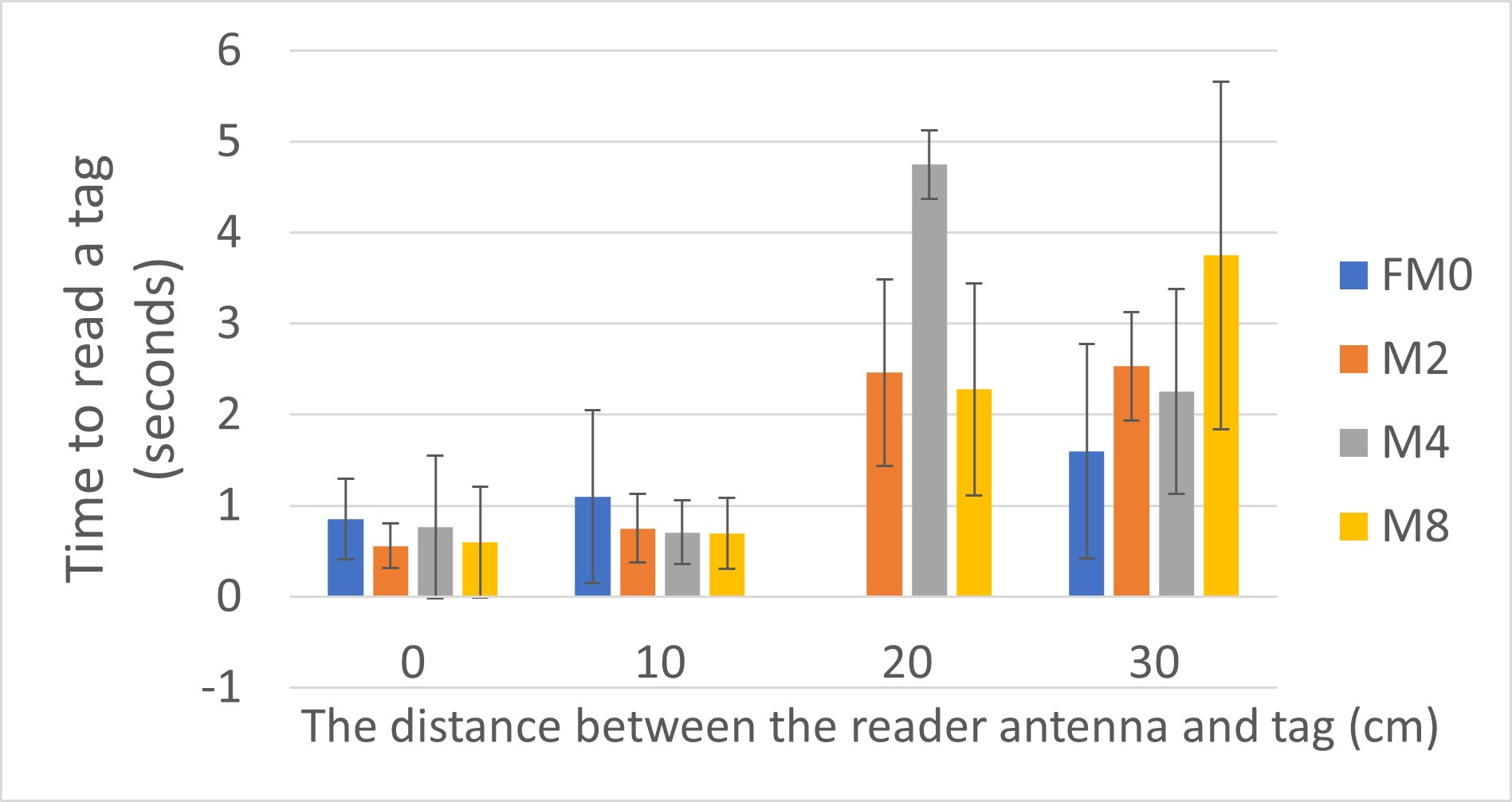}}
\caption{The duration of the inventory cycle for reading RFID sensor data.}
\label{fig:sensor_readtime}
\end{figure}

In our experiments, the reader takes less than one second to interrogate and receive sensor data from each tag when the tag-reader distance is less than 55 centimeters, as shown in Fig.~\ref{fig:sensor_readtime}. However, the time to successfully read tag sensor data can last between one and four seconds as we increase the distance to 65 centimeters, regardless of the data encoding schemes we selected. One reason lies in the less power received by the chip within each tag. Consequently, a normal inventory round can be interrupted and start over from scratch.

The results also suggest that the current design of RFID sensor tags can be improved to better support read weather sensing applications. Take the RFID-based sensor used in our experiments as an example. After receiving the EPC number from a given tag, as shown in Fig.~\ref{fig:performance_fact}, the reader needs to send a "READ" command to specify the starting location in the memory bank to read from and the number of words to be read. Only then will the tag respond with the requested data. In cases where sensing is done with direct RFID signals instead of external sensors, reading into bank memory is not necessary so extra time is not required. We present ambient-power energy harvesting as a potential solution to reduce the tag sensor reading time in Section VI-C.

%\subsubsection{RF pavement markings}

%\subsubsection{Traffic sign inventory management}

%\subsubsection{Crowdsensing for road weather conditions}

\section{Technology Gaps}
\subsection{The "GEN T" protocol for transportation}

In the EPCglobal Class 1 Generation 2 protocol, Aloha-based adaptive collision (Q-protocol) and random-number-based logical sessions are designed for singulation and flexible read rates. An inventory initiation is necessary before accessing memory banks of an RFID tag, during which the selected tag will generate a 16-bit random number (RN16), be acknowledged by the reader, and send a packet containing its protocol control bits, EPC, and CRC16 to the reader. This process could take hundreds of microseconds and could scale up quickly with Miller settings and the number of existing tags.

This time delay is not considered as a limitation for most RFID applications. However, in our study, the reader is set to be installed in the vehicle and moves with the vehicle, the time delay may have a significant impact on our results. 

The past decade has featured several studies that attempt to optimize the read rates and energy efficiency of RFID tag inventory. Su \textit{et al.} present an optimized dynamic frame slotted aloha (DFSA) anti-collision algorithm that showcases a 22-29\% speed improvement over comparable DFSA approaches in the literature \cite{su2020}. In 2023, the authors present an algorithm that attempts to resolve tag identity in a collision slot rather than discard that information outright. They demonstrate a further 83.7\% improvement in speed over their previous work \cite{su2023} as a result. Besides anti-collision algorithm improvements, the possibility of using the larger data bandwidths at frequencies such as 5.8 GHz will also contribute to read speed improvements. However, a protocol standard at these higher frequencies remains to be defined.

These efforts demonstrate the potential to improve the RFID protocol for high dynamic applications. With this motivation, a "GEN T" could be proposed for transportation systems, which is designed to minimize the inventory initiation time.

\subsection{Increasing system performance by improving antenna architecture.}
\begin{figure*}[tb!]
\centering
\includegraphics[width=0.80\textwidth]{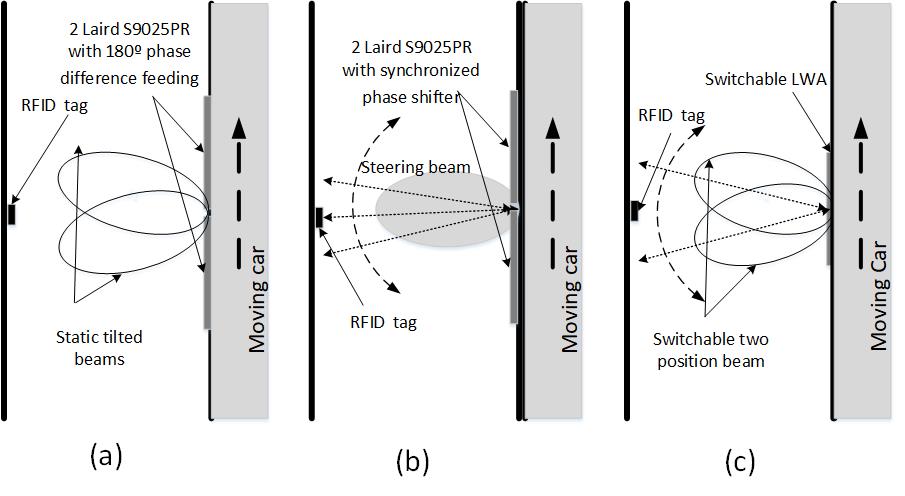}
\caption{Proposed antenna system architectures: two adjacent commercial Laird antennas out of phase (a) and with phase shifter (b), a Leaky Wave antenna ~\cite{10133315} with switching feeding port.}
\label{fig:steering beam}
\end{figure*}

It has been demonstrated that better reading performance can be achieved by setting the reader antenna at 45º. However, this solution may not be practical from aerodynamic and safety reasons, as an external protruding element is added to the car.  A tilted radiating beam could be achieved by employing a conformal antenna array configuration embedded into the car door. According to antenna array theory~\cite{Balanis-2012-antenna}, setting two antennas fed by a 3dB Wilkinson power divider and a 180º hybrid coupler added to one of the two branches will produce a radiation null along the orthogonal direction and a tilted beam. Considering a representative frequency of 902 MHz, the closest separation distance of 264 mm between the centers of two Laird S9025PR will produce a maximum array factor at an angle of 39º sideways from the orthogonal direction (Fig.~\ref{fig:steering beam}(a)). If the separation is increased, the angle will be reduced, what it is not desired. In this situation, strong mutual coupling could impact the matching of the antennas. A possible and more sophisticated work-around would consist of obtaining a steering radiating beam that could be synchronized with the speed of the car to point the target at all times (Fig.~\ref{fig:steering beam}(b)). This approach would imply the usage of a phase shifter between the two antennas instead of the previous feeding structure. However it may be unaffordable for a cost effective application due to this expensive element and the power amplifier that would make up the downside of its inherent insertion loss. A switchable beam can be an intermediate solution between a static and a continuous steering beam (Fig.~\ref{fig:steering beam}(c)). Along these lines, low-cost Leaky Wave beam scanning antennas have been recently proposed ~\cite{10133315}. Only one antenna would exhibit two beam states depending on the feeding port. The system could perform the feeding port switching operation when a decreasing in the reading rate is detected for the default position.

\subsection{Ambient power energy harvesting}
\begin{figure*}[tb!]
\centering
\includegraphics[width=0.60\textwidth]{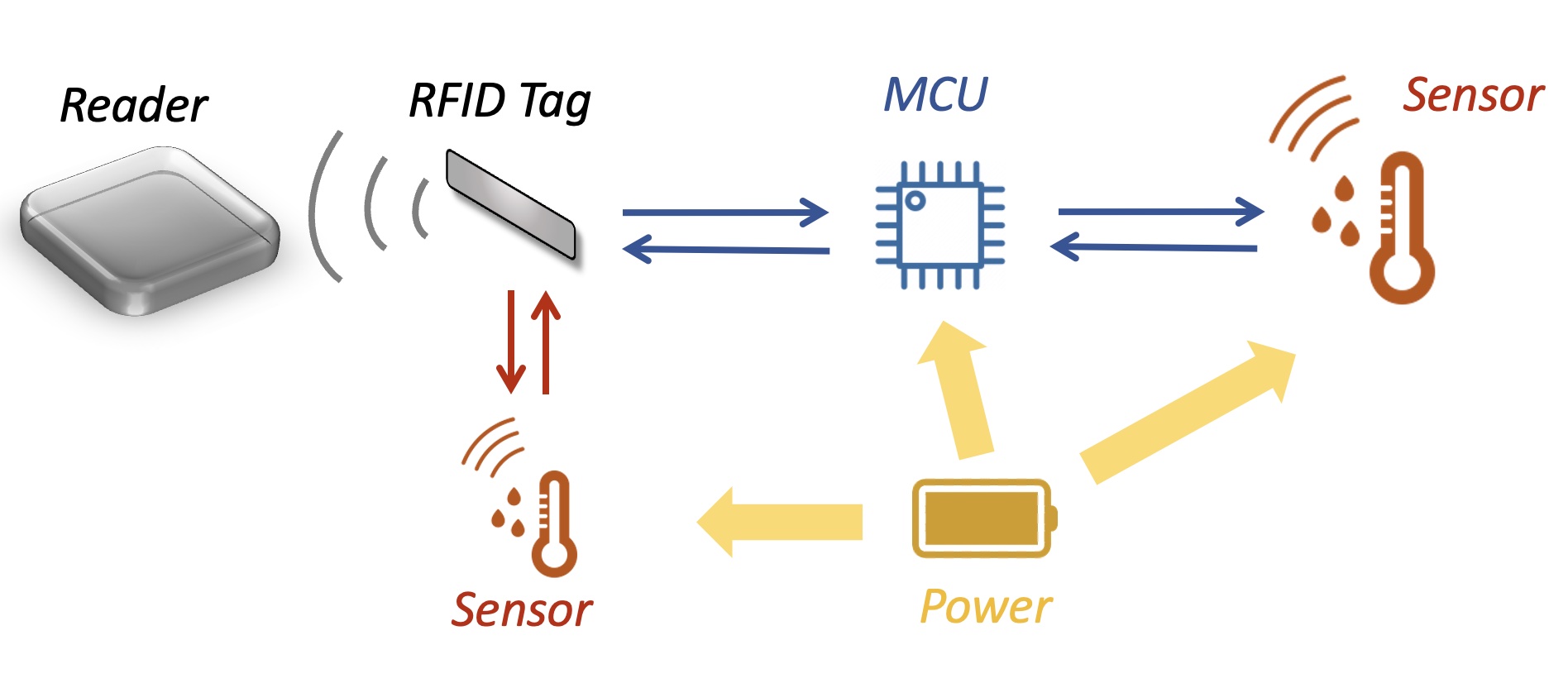}
\caption{Timing model of sensing with RFID}
\label{fig:timing}
\end{figure*}
Although REI does not require batteries for powering tags deployed on roads, it can only provide a power supply within a limited time. The on-road tags will not be able to harvest energies unless there exists passing by vehicles. This occurs when we adopt commercial RFID-based sensors whose energy supplies depend on RF signals emitted by readers. 

The discontinuous power supply from vehicle-to-tag backscattered communication can result in degraded system performance, including fewer tag reads or increased reading time. The data exchange between the on-vehicle reader and any tag requires that the strength of the RF signal received by the chip within the tag is greater than a predetermined threshold. Otherwise, the inventory round will be interrupted and start over again from scratch, reducing the likelihood of successful reading, as shown in Fig.~\ref{fig:performance_fact}. Additionally, when an RFID tag is used for environmental monitoring, extra power is needed to activate the sensing modules embedded within the tag, which may incur extra reading time.

Fortunately, the advancement in ambient backscatter communication~\cite{torres2021backscatter} has made possible pervasive energy harvesting from various forms of commercial radio signals. In addition to RF signals from on-vehicle readers, tags can harvest power from radio waves emitted by cellular stations, WiFi routers~\cite{kellogg2014wi}, Frequency Modulated (FM) towers~\cite{daskalakis2017ambient}, and television centers~\cite{liu2013ambient}.
 
RFID tags could operate in an active mode, where external power sources like batteries or power harvesting circuits are connected to the tag's IC.  The activation time of these components reduces significantly and thus broadcasting the sensor information becomes possible. 

To enhance RFID sensor capabilities, integrating a microcontroller into the system and linking the sensor to it, rather than directly embedding it in the tag IC, is a viable approach. Sensor data can be transmitted to the tag using binary sequences through interfaces like GPIO, SPI, or I2C. However, this process may introduce additional time delays due to the round-trip transmission of sensor data between the microcontroller and the tag IC.

Upon a more detailed examination of temporal aspects, the comprehensive duration for acquiring sensor data via RFID tag interrogation can be succinctly delineated by this overarching model Fig. \ref{fig:timing}:  

\begin{equation}
\begin{aligned}
\text{T\_total} &= \text{Query processing time} \\ &+ \text{T\_activation (Tag IC , Sensor, MCU)} \\
&+ \text{Propagation time} 
\end{aligned}
\end{equation}

Herein, Query processing time pertains to the interval dedicated to RFID tag reading; T\_activation encapsulates the temporal expenditure associated with activating the Tag IC alone, or the combined activation of Tag IC and Sensor, or encompassing Tag IC, Sensor, and Microcontroller, contingent upon the presence of a Microcontroller and external power sources; Propagation time denotes the temporal interval encompassing the transmission of bit streams back to the RFID IC when a microprocessor is incorporated, encompassing potential additional time for accessing auxiliary memory banks, such as the User Memory bank, during the readout process.

\section{Conclusion}
This study presented a comprehensive exploration of the REI system, specifically emphasizing the effects of the EPC UHF GEN 2 protocol on its performance in real-world urban environments. The primary goal was to elucidate the factors influencing the efficiency of the RFID-enhanced transportation infrastructure, given the rising demand for intelligent road systems.

Our results reveal significant dependencies of REI's effectiveness on various parameters:

\begin{itemize}
\item The encoding scheme plays a pivotal role in optimizing REI's performance, with the Miller 2 encoding scheme emerging as the most superior, balancing throughput and noise cancellation effectively.
\item Vehicle speed was identified as a factor, especially in RF lane marking, where degradation in performance becomes apparent at speeds exceeding 40 mph.
\item The influence of road geometry-induced path loss has been dissected, emphasizing the inverse relationship between tag-reader distance and the number of readings.
\item Tag installation height has also been recognized as crucial, revealing the potential of harnessing the multi-path effect to boost reader-tag communication.
\item On-vehicle antenna orientation has a pronounced impact on performance, suggesting the need for optimal antenna designs that consider both reader coverage and path loss.
\item The study also shed light on the intricacies of interrogating user memory banks in RFID sensor tags, which could be instrumental in weather-sensing applications.
\end{itemize}

While the findings provide constructive insights into the promising prospects of REI in transforming urban transportation systems, certain limitations were also brought to light. Specifically, while REI meets the demands for traffic sign inventory management and environmental monitoring, it falls short in catering to high-speed navigation, emphasizing the need for further optimization or support by new protocols (e.g., GEN T).

The broader implications of this research suggest a path towards a refined and efficient REI system capable of elevating the intelligence of road infrastructures. We envision this research as a cornerstone for urban planners, technologists, and even policymakers alike as they navigate the complexities of transportation infrastructure with increased connectivity.

%\bibliographystyle{IEEEtran}
%\bibliography{bib.bib}
\printbibliography

\newpage

%\section{Biography Section}

\vspace{11pt}
\vskip -2\baselineskip plus -1fil
%\bf{If you include a photo:}\vspace{-33pt}
\begin{IEEEbiography}[{\includegraphics[width=1in,height=1.25in,clip,keepaspectratio]{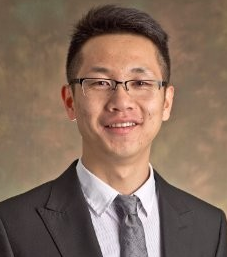}}]{Dajiang Suo}  is an Assistant Professor at Arizona State University. Prior to that, he served as a research scientist at Massachusetts Institute of Technology. Suo obtained a Ph.D. in Mechanical Engineering from MIT in 2020. Suo holds a B.S. degree in Mechatronics Engineering, and an S.M. degree in Computer Science and Engineering Systems. His research interests include the Internet of Things, connected vehicles, cybersecurity, and RFID.

Before returning to school to pursue PhD degree, Suo was with the vehicle control and autonomous driving team at Ford Motor Company (Dearborn, MI), working on the safety and cyber-security of automated vehicles. He also serves on the Standing Committee on Enterprise, Systems, and Cyber Resilience (AMR40) at the Transportation Research Board.
\end{IEEEbiography}

\vspace{11pt}

%\bf{If you will not include a photo:}\vspace{-33pt}
\begin{IEEEbiography}[{\includegraphics[width=1in,height=5in,clip,keepaspectratio]{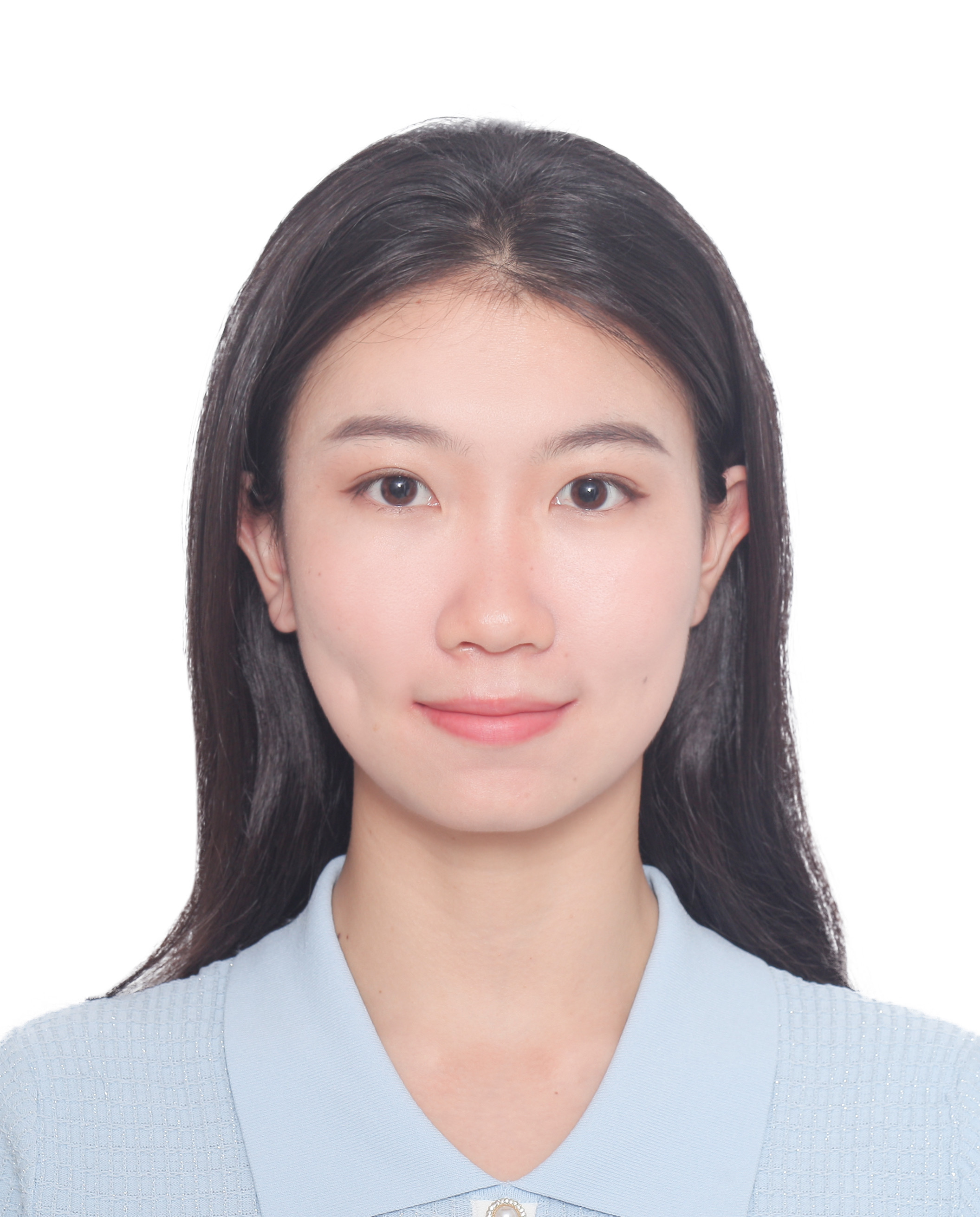}}]{Heyi Li} is a graduate student in Mechanical Engineering at the Auto ID Labs, Massachusetts Institute of Technology, Cambridge, MA, USA. She received the B.S. degree in Microelectronics Science and Engineering from Southern University of Science and Technology (SUSTech) in 2021 and S.M. in Mechanical Engineering from Massachusetts Institute of Technology in 2023. Her research interests include RFID, Internet of Things (IoT), wireless network and smart sensing.

\end{IEEEbiography}

\vspace{11pt}

%\bf{If you will not include a photo:}\vspace{-33pt}
\begin{IEEEbiography}[{\includegraphics[width=1in,height=1.25in,clip,keepaspectratio]{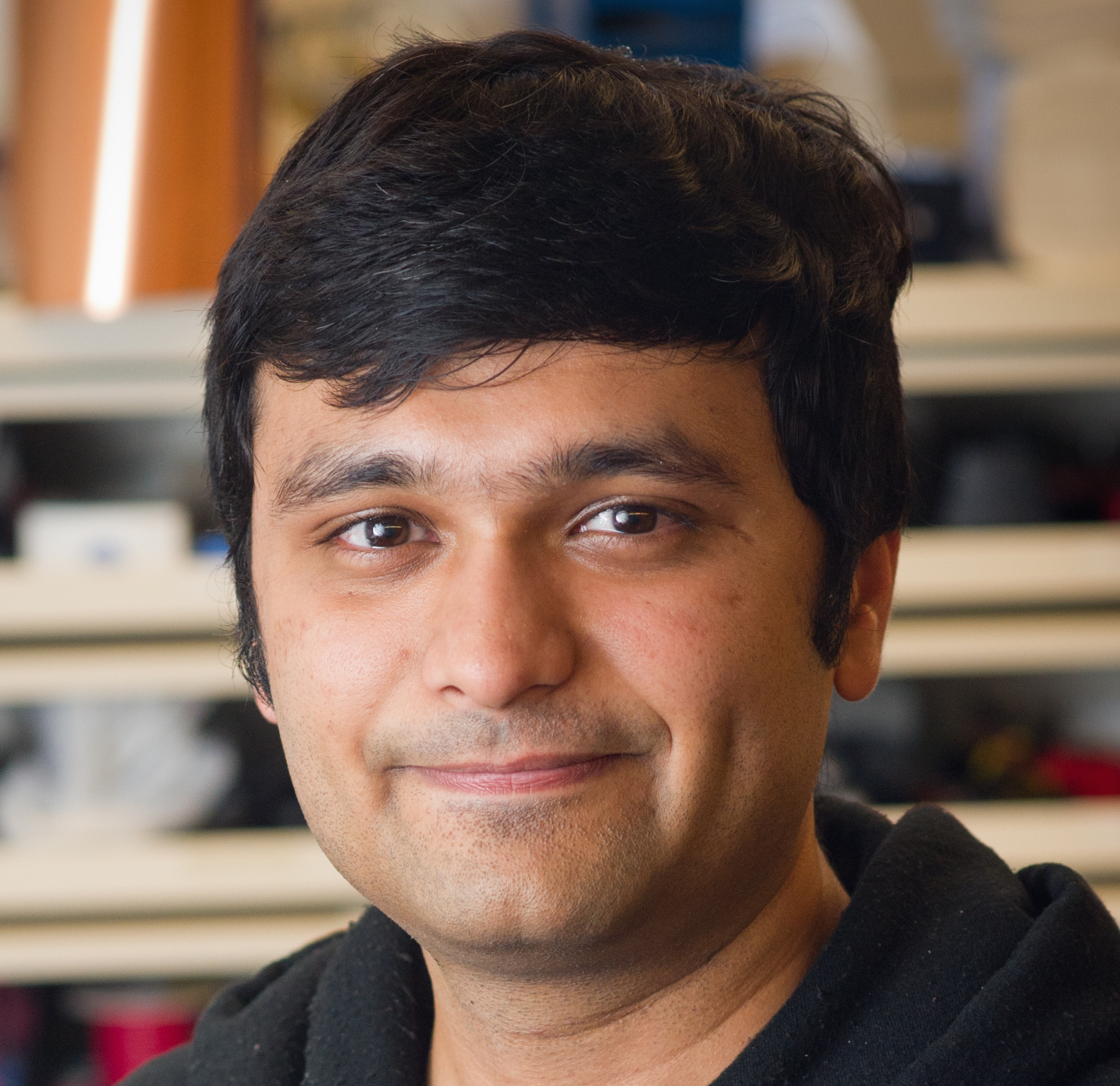}}]{Rahul Bhattacharyya} Rahul Bhattacharyya received the Ph.D.
degree in Systems Engineering from the
Massachusetts Institute of Technology,
Cambridge, MA, USA, in 2012. He is
currently the Director of the Auto-
ID Labs at MIT, where his research
encompasses the development and
integration of technologies that form the
framework for the Internet of Things. He is an
Associate Editor for the IEEE Journal of RFID and IEEE Sensors.
He has also served as the TPC Chair of
IEEE RFID 2018-19 and the 4th International Conference on the
Internet of Things (IoT 2014) at MIT. He was also Guest Editor for a
special issue on the Internet of Things for the IEEE Transactions on
Automotive Science and Engineering. He has authored over 50
technical peer reviewed publications. He was the recipient of the 2020
IEEE Sensors and 2015 IEEE RFID-TA Conference Best Paper
Award.
\end{IEEEbiography}

\vspace{11pt}

%\bf{If you will not include a photo:}\vspace{-33pt}
\begin{IEEEbiography}[{\includegraphics[width=1in,height=1.25in,clip,keepaspectratio]{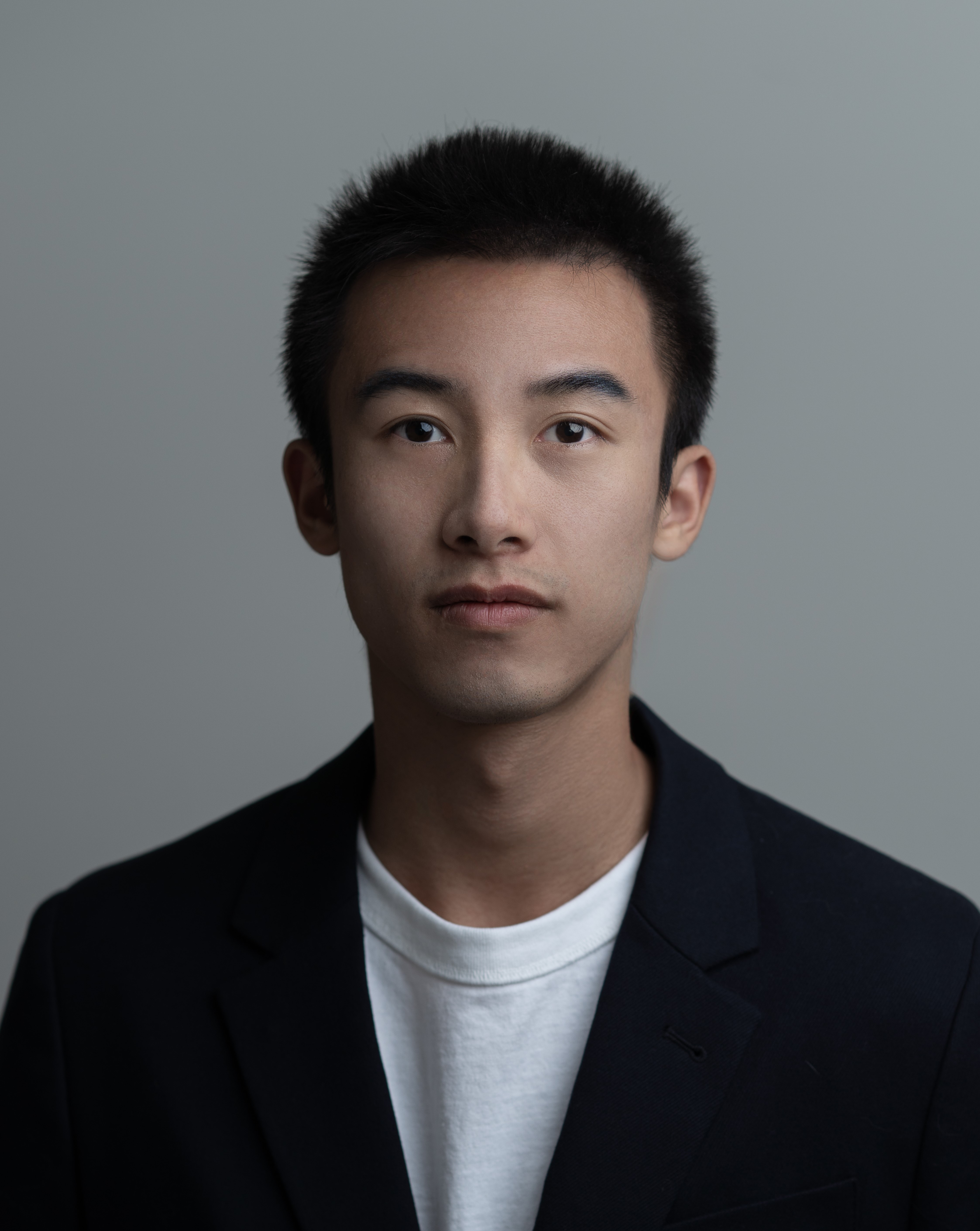}}]{Zijin Wang} is currently a PhD candidate at the Department of Civil, Environmental, and Construction Engineering, University of Central Florida, Orlando, FL, USA. He received the B.S. degree in logistics engineering from Central South University, Changsha, in 2020. His research interests include traffic safety analysis, intelligent transportation system, connected and automated vehicles, microscopic traffic simulation, co-simulation, and digital twin.
\end{IEEEbiography}

\vspace{11pt}

%\bf{If you will not include a photo:}\vspace{-33pt}
\begin{IEEEbiography}[{\includegraphics[width=1in,height=1.25in,clip,keepaspectratio]{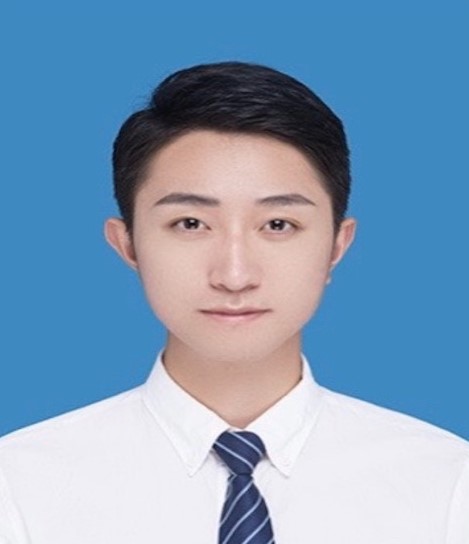}}]{Shengxuan Ding} is a Ph.D. student at the Department of Civil, Environmental, and Construction Engineering, University of Central Florida, Orlando, FL, USA. He received his M.S. and B.S. degrees in
transportation engineering from Chang'an University. His research interests include traffic safety analysis, intelligent transportation systems, and connected and automated vehicles.
\end{IEEEbiography}

\vspace{11pt}

%\bf{If you will not include a photo:}\vspace{-33pt}
\begin{IEEEbiography}[{\includegraphics[width=1in,height=1.25in,clip,keepaspectratio]{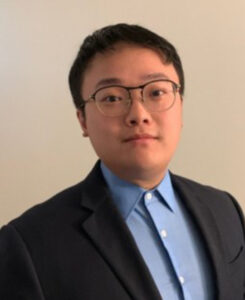}}]{Ou Zheng} was the lead research engineer at the Smart and Safe Transportation Lab, at the University of Central Florida, whose research interests center on the application of computer vision, digital twins, and large language model techniques to enhance traffic safety. He has been instrumental in numerous high-impact projects, including the UCF SST Real-time crash risk visualization platform, which won the US Department of Transportation Solving for Safety Visualization Challenge and became one of six USDOT Safety Data Initiative tools. Dr. Zheng has also created CitySim, an open-source dataset of drone video vehicle trajectory and digital twin environment, which has gained significant recognition from various research fields worldwide.
\end{IEEEbiography}

\vspace{11pt}

%\bf{If you will not include a photo:}\vspace{-33pt}
\begin{IEEEbiography}[{\includegraphics[width=1in,height=1.25in,clip,keepaspectratio]{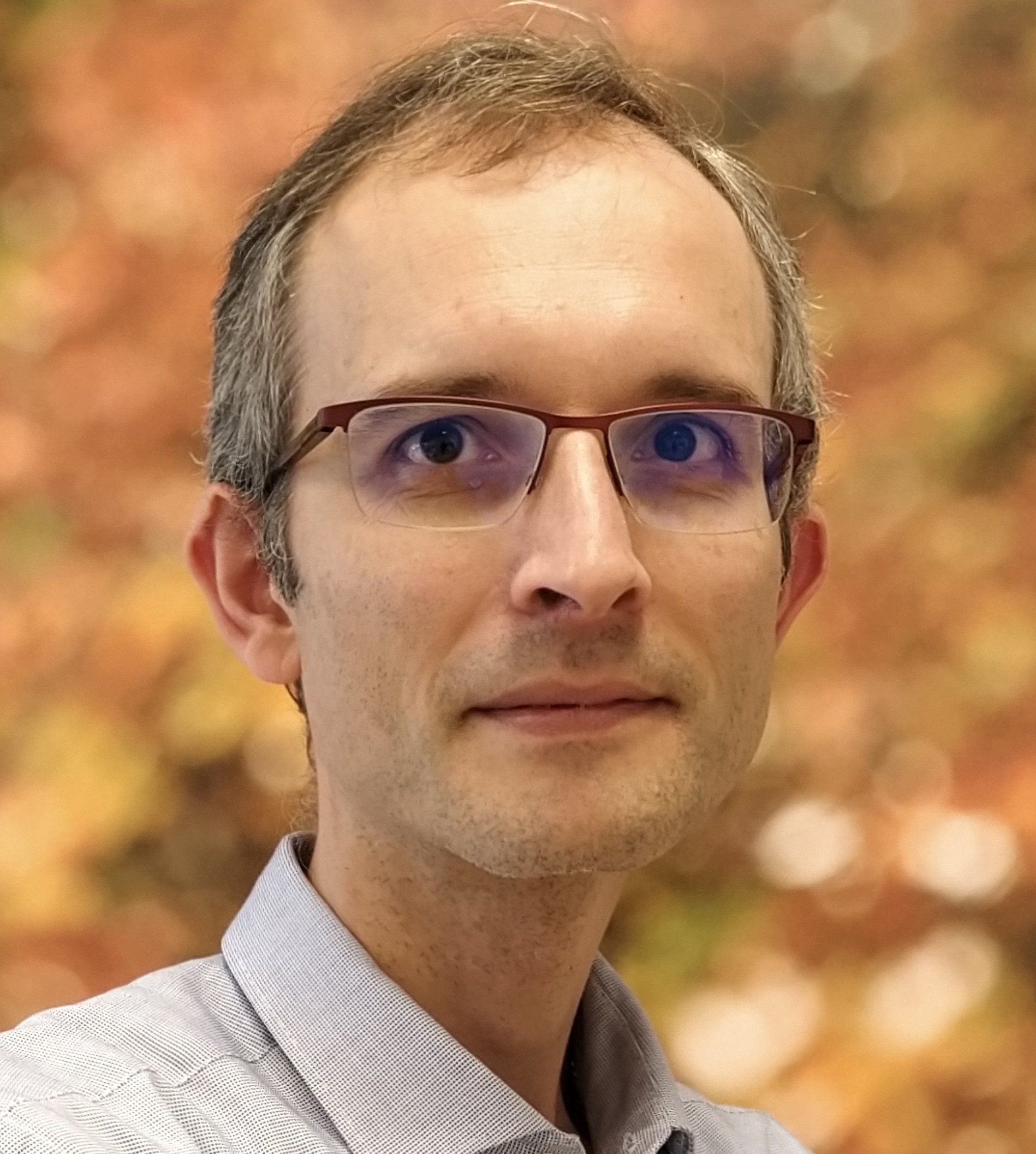}}]{Daniel Valderas}received the MSc. degree and the PhD from Tecnun, University of Navarra, in 1998 and 2006 respectively. He was Academic Visitor at Florida Atlantic University in 2003. He joined Ceit-IK4 as Research Scientist and Tecnun as Assistant Lecturer in 2006. He was visiting researcher at Imperial College in 2007, at Queen Mary, University of London, in 2008 and at MIT in 2022. He has participated in many different research projects on the Antenna and Electromagnetics with practical background in Electromagnetic Simulation. His research interests focus on UWB and broadband antennas, RFID and implanted antennas, wireless chipless sensors and radar. Starting in 2018, he is currently Associate Professor and Research Scientist with Tecnun, University of Navarra.
\end{IEEEbiography}

\vspace{11pt}

%\bf{If you will not include a photo:}\vspace{-33pt}
\begin{IEEEbiography}[{\includegraphics[width=1in,height=1.25in,clip,keepaspectratio]{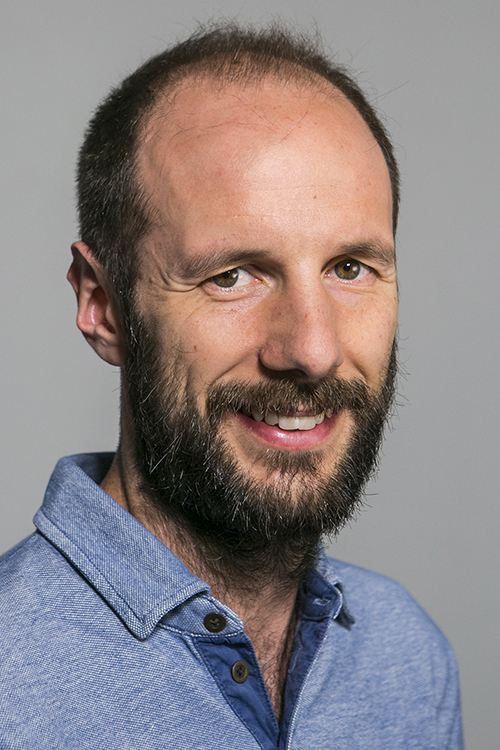}}]{Joan Melià-Seguí} (Senior Member, IEEE) received the B.Sc. and M.Sc. degrees in telecommunications engineering from the Universitat Politècnica de Catalunya and the Ph.D. degree from the Universitat Oberta de Catalunya (UOC). He has been a Researcher at the Universitat Pompeu Fabra, Visiting Researcher at the Palo Alto Research Centre (Xerox PARC), and Fulbright Visiting Scholar at the Massachusetts Institute of Technology. He is currently Associate Professor at the Faculty of Computer Science, Multimedia and Telecommunication Engineering, UOC. He has authored over 40 technical peer-reviewed publications. His research interests include sustainability and low-cost RF sensing.
\end{IEEEbiography}

\vspace{11pt}

%\bf{If you will not include a photo:}\vspace{-33pt}
\begin{IEEEbiography}[{\includegraphics[width=1in,height=1.25in,clip,keepaspectratio]{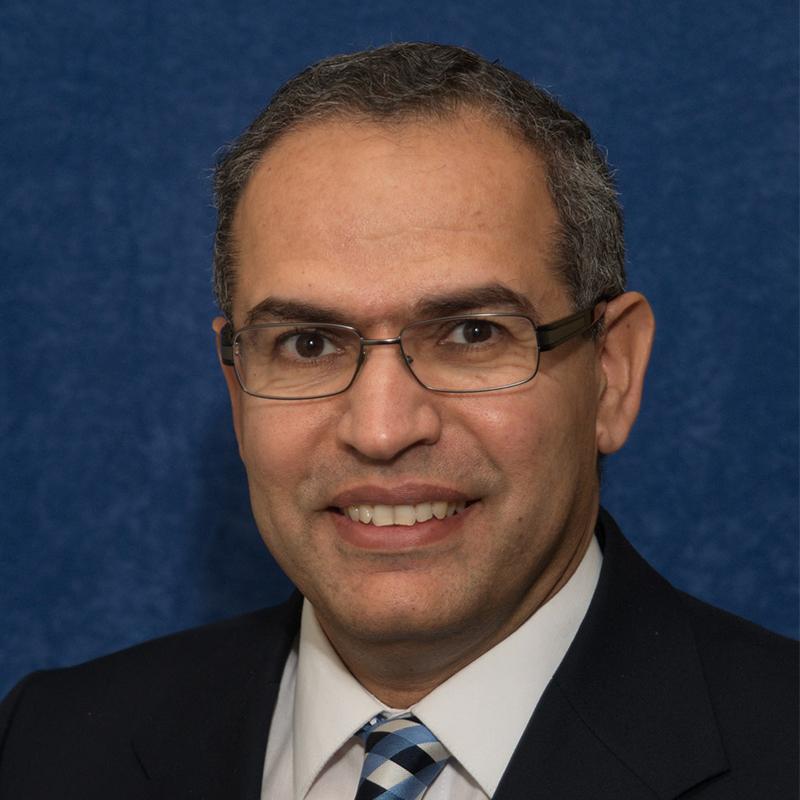}}]{Mohamed Abdel-Aty} is a Trustee Endowed Chair at the University of Central Florida (UCF). He is a Pegasus Distinguished Professor and former Chair of the Civil, Environmental and Construction Engineering Department at UCF. He has a joint appointment with the department of Computer Science. He is the founder and leading the Future City initiative at UCF. As part of the initiative, he introduced in 2019 the first MS degree in Smart Cities in Engineering in the US. His main expertise and interests are in the areas of traffic safety, transportation technology, simulation, big data and data analytics, ITS and CAV. He is pioneer and well recognized in research in real-time safety, Proactive traffic management, and Connected Vehicles. In 2015, he was awarded the Pegasus Professorship, the highest honor at UCF. Dr. Abdel-Aty has managed over 80 research projects. Dr. Abdel-Aty has published more than 800 papers, 425 in journals. He is a member of the Academy of Science, Engineering and Medicine of Florida (ASEMFL).  He is a Fellow of both ASCE and ITE, and Senior member of IEEE. He is the former Chair of the ASCE Transportation Safety Committee. Dr. Aty has received the 2020 Roy W. Crum Distinguished Service Award from the Transportation Research Board, National Safety Council’s Distinguished Service to Safety Award, Francis C. Turner award from ASCE, the 2019 Transportation Safety Council Edmund R. Ricker Award, Institute of Transportation Engineering (ITE) and the Lifetime Achievement Safety Award and S.S. Steinberg Award from ARTBA in 2019 and 2022, respectively. He has also received multiple International awards including the Prince Michael International Road Safety Award, London 2019. He has delivered more than 30 Keynote speeches in conferences around the world. He is a registered professional engineer in Florida.
\end{IEEEbiography}

\vspace{11pt}

%\bf{If you will not include a photo:}\vspace{-33pt}
\begin{IEEEbiography}[{\includegraphics[width=1in,height=1.25in,clip,keepaspectratio]{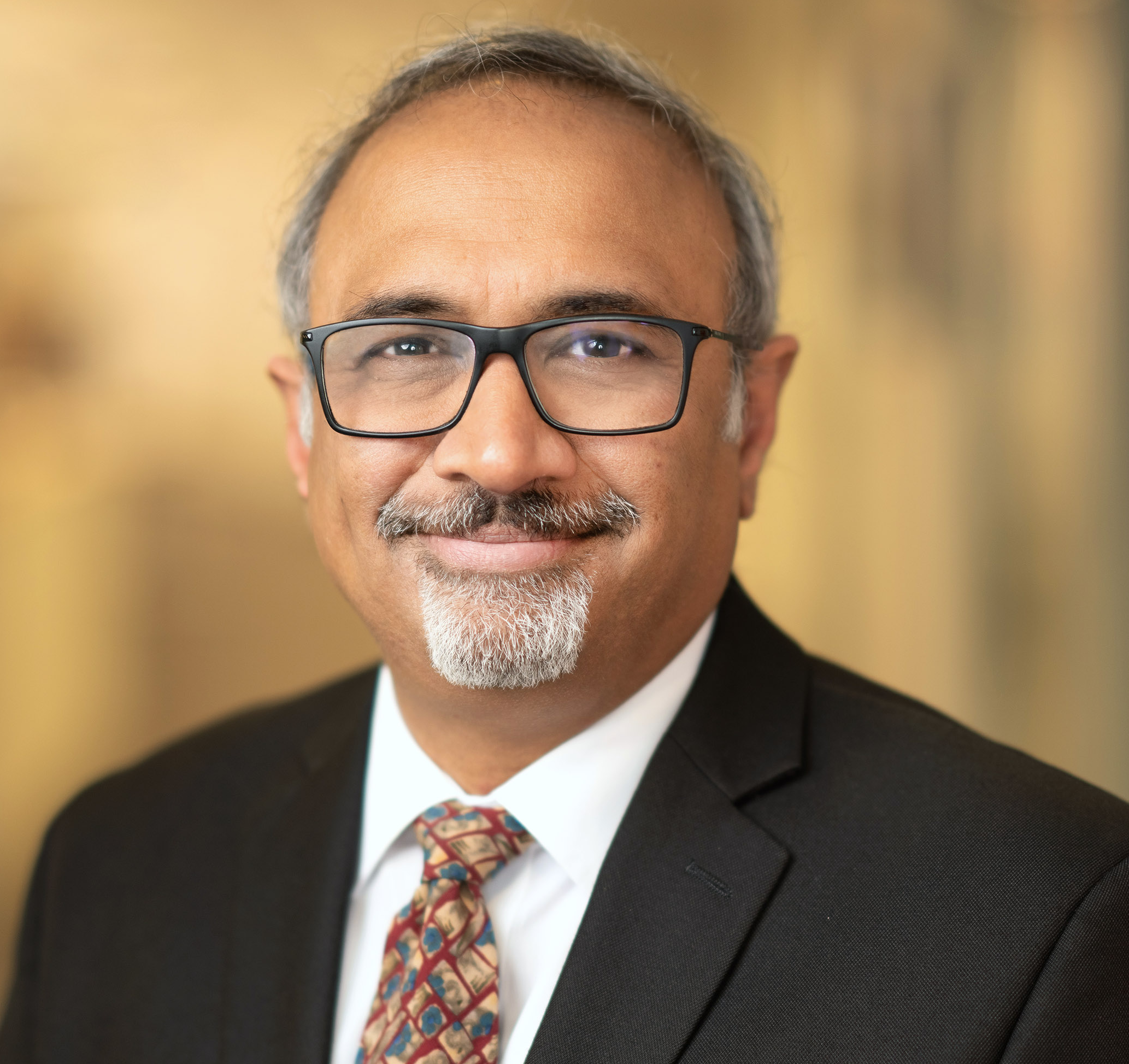}}]{Sanjay E. Sarma} is the Fred Fort Flowers (1941) and Daniel Fort Flowers (1941) Professor of Mechanical Engineering at MIT. He co-founded the Auto-ID Center at MIT and developed many of the key technologies behind the EPC suite of RFID standards now used worldwide. He was also the founder and CTO of OATSystems, which was acquired by Checkpoint Systems (NYSE: CKP) in 2008. He serves on the boards of GS1, EPCglobal and several companies including CleanLab and Aclara Resources (TSX:ARA). 

Dr. Sarma received his Bachelors from the Indian Institute of Technology, his Masters from Carnegie Mellon University and his PhD from the University of California at Berkeley. Sarma also worked at Schlumberger Oilfield Services in Aberdeen, UK. He has authored over 150 academic papers in computational geometry, sensing, RFID, automation and CAD, and is the recipient of numerous awards for teaching and research including the MacVicar Fellowship, the Business Week eBiz Award and Informationweek’s Innovators and Influencers Award.
\end{IEEEbiography}

%\section{Appendix}

%\begin{table*}[]
%\centering
%\begin{tabular}{ccccc}
%\hline \multirow{$x_a$ (m)} & \multicolumn{4}{c}{d (m)} \\
%\cline {2 - 5} &  8.5 (L1$\vert$L2$\vert$L3) & 16.0 (L1$\vert$L2$\vert$L3) & 23.5 (L1$\vert$L2$\vert$L3) & 31.0 (L1$\vert$L2$\vert$L3)\cr
%\hline
%-206.0 & 0$\vert$0$\vert$0 & - & - & - \cr
%-198.5 & 0$\vert$0$\vert$0 & 0$\vert$0$\vert$0 & 0$\vert$0$\vert$0 & - \cr
%-191.0 & 0$\vert$0$\vert$0 & 0$\vert$0$\vert$0 & 0$\vert$0$\vert$0 & - \cr 
%1.0 & 0$\vert$0$\vert$0 & 0$\vert$0$\vert$0 & 0$\vert$0$\vert$0 & 0$\vert$0$\vert$0 \cr 
%8.5 & 0$\vert$0$\vert$0 & 0$\vert$0$\vert$0 & 0$\vert$0$\vert$0 & 0$\vert$0$\vert$0  \cr 
%16.0 & - & 0$\vert$0$\vert$0 & 0$\vert$0$\vert$0 & 0$\vert$0$\vert$0 \cr 
%23.5 & - & - & - & 0$\vert$0$\vert$0 \cr 
%\hline
%\end{tabular}
%\caption{Mixed autonomy jordan controller throughput comparison as a function of EMS vehicle maximum speed. Here $L1, L2$ and $L3$ denote the varying maximum speeds of EMS vehicle and are defined to be $L1= 30m/s$, $L2= 35 m/s$ and $L3=40 m/s$.}
%\label{compare-q2}
%\end{table*}

\vfill

\end{document}